\documentclass{article}

\usepackage[preprint]{corl_2026} 

\usepackage{algorithm}
\usepackage{algpseudocode}
\usepackage{float}   
\usepackage{hyperref}       
\usepackage{url}            
\usepackage{booktabs}       
\usepackage{amsfonts}       
\usepackage{nicefrac}       
\usepackage{microtype}      
\usepackage{xcolor}         
\usepackage{enumitem}
\usepackage{graphicx}
\usepackage{subcaption}
\usepackage{amsmath}

\usepackage{amsthm}
\newtheoremstyle{tight}%
  {0.3em}
  {0.3em}
  {\itshape}
  {}
  {\bfseries}
  {.}
  {0.5em}
  {}
\theoremstyle{tight}
\newtheorem{proposition}{Proposition}

\newcommand{\SE}{\mathrm{SE}}

\newcommand{\ours}{\text{Pix2Act}}
\usepackage{caption}
\usepackage{titlesec}
\titlespacing*{\section}
{0pt}{0.2em}{0.1em}

\titlespacing*{\subsection}
{0pt}{0.04em}{0.025em}

\usepackage[belowfloats]{footmisc}
\usepackage{amssymb}
\usepackage{wrapfig}
\usepackage{subcaption}
\usepackage{multirow}
\usepackage{xcolor}

\usepackage{graphicx}
\definecolor{darkblue}{RGB}{0, 70, 140}
\title{Pix2Act: Image-Space Manipulation Policies with Equivariant Augmentation}

\author{
  \textbf{Haojie Huang}$^{1,\ast}$ \quad
  \textbf{Linfeng Zhao}$^{2}$ \quad
  \textbf{Haotian Liu}$^{1}$ \quad
  \textbf{Zhang Ye}$^{1}$ \\[-0.05em]
  \textbf{Si-Yuan Huang}$^{3}$ \quad
  \textbf{Mingxi Jia}$^{4}$ \quad
  \textbf{Boce Hu}$^{1}$ \quad
  \textbf{Fangzhou Lin}$^{5}$ \quad
  \textbf{Yu Qi}$^{1}$ \\[-0.05em]
  \textbf{Dian Wang}$^{2}$ \quad
  \textbf{Robin Walters}$^{1,\dagger}$ \quad
  \textbf{Robert Platt}$^{1,\dagger}$ \\[-0.05em]
  $^{1}$Northeastern University
  \quad
  $^{2}$Stanford University
  \quad
  $^{3}$University of Pennsylvania \\[-0.05em]
  $^{4}$Brown University \quad
  $^{5}$Texas A\&M University \quad
  $^{\ast}$Corresponding author
  \quad
  $^{\dagger}$Equal advising \\[-0.05em]
  \href{https://haojhuang.github.io/pix2act_page/}{\texttt{haojhuang.github.io/pix2act\_page}}
  \hfill
  \texttt{huang.haoj@northeastern.edu}
}

\begin{document}
\maketitle


\begin{abstract}
Representing manipulation actions as 2D trajectories in the camera plane provides a compact and interpretable basis for learning complex 3D manipulation policies. However, it also creates challenges from out-of-frame trajectories and limited precision. We propose $\ours$, an imitation learning method that addresses these challenges by generating \textit{continuous} image-space keypoint trajectories in each camera plane and losslessly recovering end-effector poses via triangulation.
This reformulates high-dimensional 3D control as a simpler, more learnable 2D prediction problem. Crucially, it aligns observations and actions in the same coordinate space, enabling equivariant transformations to jointly rotate individual camera images together with their image-space actions. We analyze the symmetry properties of this augmentation and design a network architecture that can fuse multiple camera views while respecting their per-view rotations. As a result, $\ours$ implicitly enlarges the support of the data distribution and learns invariant action structures across transformations, yielding improved generalization and overall performance. Across diverse simulated and real-world manipulation tasks, $\ours$ outperforms state-of-the-art baselines and remains robust under camera perturbations.
\end{abstract}

\section{Introduction}

Manipulation policy learning has seen significant advances, particularly in mimicking 3D trajectories for closed-loop control. Typically, prior work models a conditional mapping from 2D visual observations to a 3D action chunk, defined as a sequence of 3D translations and rotations. However, this formulation often treats the image space and the 3D Cartesian action space as independent domains, overlooking the inherent geometric relationship between them. This unstructured mapping lacks explicit spatial grounding and forces the network to implicitly infer complex 3D trajectories between the end-effector and objects, making the learning problem highly ambiguous and prone to overfitting. Projecting manipulation actions as 2D trajectories in the camera plane offers an alternative: complex 3D policies can be learned through compact and interpretable 2D representations. However, this is non-trivial due to precision limits and the difficulty of cross-view fusion. For example, prior work~\cite{ren2025motion} denoises pixel keypoint coordinates independently in each view and triangulates them across two agent cameras to recover 3D positions. As a result, it suffers from precision loss due to pixel discretization, an inability to represent actions outside the image frame, large triangulation errors under widely separated viewpoints, and cross-view tracking inconsistencies.

\begin{figure}[t]
    \centering
    \includegraphics[width=0.95\linewidth]{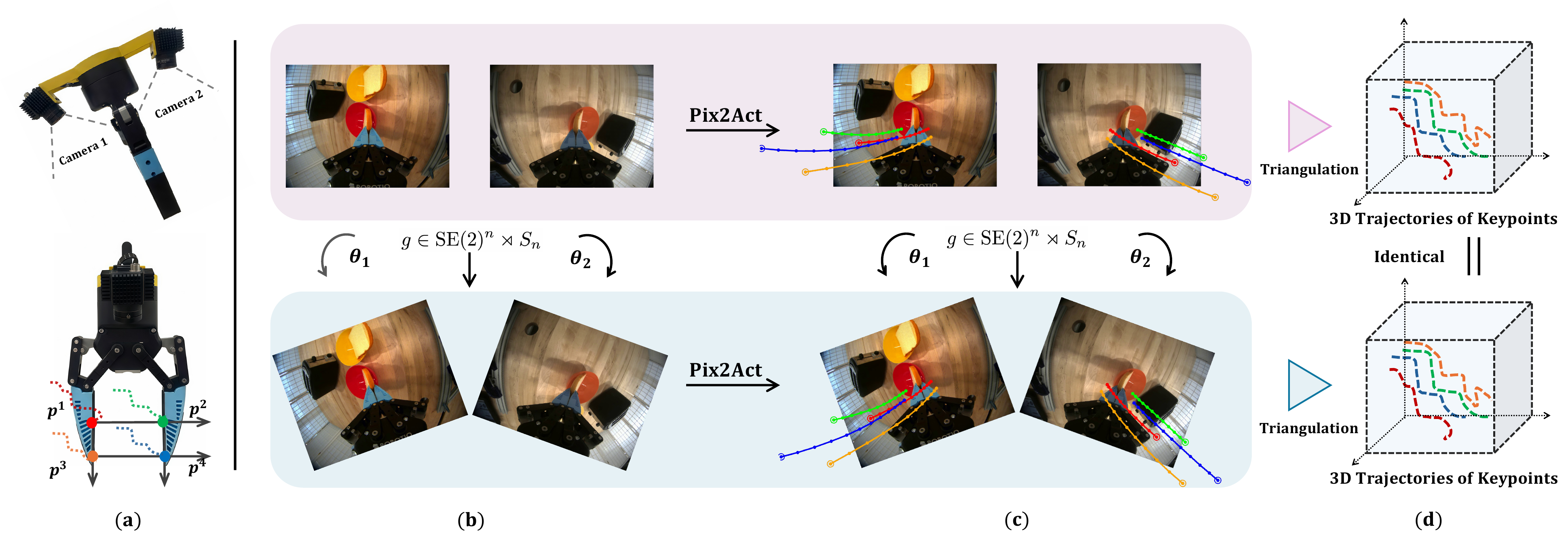}
    \caption{From left to right: \textbf{(a)} an illustration of our dual in-hand camera setup and the selected keypoints; \textbf{(b,c)} our proposed method consumes stereo images and predicts image actions for these keypoints; independent image rotations result in consistent transformations of the 2D action trajectories; \textbf{(d)} triangulation yields consistent 3D results for both the original and transformed image pairs, as they convey the same contextual information.}
    
    \label{fig:action_repr}
\end{figure}

In this work, we provide the policy with explicit spatial grounding by shifting the action representation from trajectories in $\SE(3)$ to a set of continuous and unbounded paths in the image plane, i.e. in $\mathbb{R}^2$. This formulation eliminates the pixel discretization that limits prior 2D approaches, recovering full precision in the action representation.
Figure~\ref{fig:action_repr} illustrates the key elements of our approach. First, we define a set of keypoints on the gripper (lower left of Figure~\ref{fig:action_repr}a). We encode a trajectory through $\SE(3)$ (the action chunk) with trajectories of these keypoints. Second, we project these keypoints into image space to obtain 2D supervision, and train a diffusion model to generate their trajectories in two stereo image planes (Figure~\ref{fig:action_repr}bc). Finally, we reconstruct the full action chunk in 3D through triangulation from the two stereo images into $\mathbb{R}^3$ (Figure~\ref{fig:action_repr}d). 
By leveraging camera geometry, this projection–triangulation pipeline is fully invertible, establishing a tight geometric coupling between observation and action while keeping the shift from $\SE(3)$ to image-space paths lossless.

Since the actions and observations reside in the same 2D image space,
we can achieve equivariant action augmentation using purely image-space transformations. As shown in Figure~\ref{fig:action_repr}bc, if one of the observed images rotates, the corresponding generated action trajectories in that image should be transformed by the same rotation. Importantly, this data augmentation is carried out independently for each of the two camera images. We call the symmetry associated with this data augmentation \textit{$n$-equivariance} where $n$ is the number of cameras (2 in this case). 
It enables us to encode a useful symmetry into the generative action model -- that the generated action trajectories in the image plane should be equivariant to independent rotations of each camera around its optical axis. 
This data augmentation is important because it biases the model toward responding to local image features, as the policy needs to steer the end effector to manipulate objects under different transformations using the predicted image trajectories.
Besides, it also enlarges the distribution of data used to train the model 
to learn a smoother function over an effectively lower dimensional space~\cite{song2019generative, song2020score,alain2014regularized}. 

Our approach is compatible with any pair of cameras with intersecting optical axes and we focus on the case where the two cameras are mounted near the end effector (Figure~\ref{fig:action_repr}a top). 
This setup offers several advantages.
First, the close proximity of the in-hand cameras to the end-effector enables our model to see the objects being manipulated with high acuity, providing clearer patterns of how the end-effector tips interact with the target object. 
Second, this configuration is readily deployable and can be easily integrated with different robotic arms.
Finally, projecting the action chunk onto the in-hand camera views naturally canonicalizes the action with respect to the end-effector frame. As a result, the representation becomes inherently SE(3)-invariant, substantially improving policy generalization~\cite{wang2025practical}.
We refer to our proposed approach as $\ours$ for simplicity, and summarize our contributions as follows:

\begin{enumerate}[label=(\arabic*),itemsep=1pt, topsep=0pt, parsep=0pt, partopsep=0pt]
    \item We propose a new action representation that expresses 3D action chunks using \textit{continuous} and \textit{unbounded} 2D image coordinates relative to a pair of in-hand cameras;
    \item We propose an equivariant data augmentation technique for image-based action learning, and introduce the concept of \textit{$n$-equivariance} to formalize its underlying symmetry properties, leading to improved policy learning performance.
    \item We propose Diffusion X-Net, a novel neural network architecture that uses a multi-view transformer to fuse inputs from different cameras, together with a DiT diffusion head, making it well suited for learning \textit{$n$-equivariant} functions.
    \item We demonstrate that our method outperforms several strong baselines across a range of simulation and real-world tasks.
\end{enumerate}

\section{Related Work}

\textbf{Data augmentation.} Data augmentation is widely used in vision, self-supervised learning, and language model training to thicken the data manifold and improve generalization~\cite{simard2003best, laskin2020reinforcement, radford2021learning, li2022blip, kafle2017data, xiao2024florence, wei2022chain, shorten2021text, sennrich2016improving, wei2021finetuned}. For robotic policy learning, we distinguish two types. \textit{Invariant action augmentation} perturbs observations (Gaussian noise, random crops, lighting and background variations) while keeping actions fixed, encouraging robustness to nuisance variation~\cite{tobin2017domain, dasari2019robonet, fang2023anygrasp, jiang2021synergies, robomimic2021, hansen2020self, laskin2020reinforcement}. \textit{Equivariant action augmentation} jointly transforms observations and actions to synthesize new training pairs, but requires the observation and action spaces to be aligned. Action-centric representations achieve this by modeling actions as distributions over pixel or voxel spaces, where transformations of observations induce corresponding transformations of actions~\cite{shridhar2022cliport, zeng2021transporter, wang2020policy, shridhar2023perceiver, goyal2023rvt, goyal2024rvt, huang2022equivariant}. Several works extend this idea to closed-loop settings: \citet{jia2022seil} for $\SE(2)$ policies, \citet{wang2024equivariant} using voxel representations for 3D actions, and \citet{hu20253d} via lifting 2D images to a sphere. Unlike these, we project 3D action chunks onto continuous, unbounded image coordinates, enabling equivariant augmentation for 3D closed-loop policy learning.

\textbf{Robotic action representation.} Action representation fundamentally shapes the learning problem. Most manipulation methods parameterize actions in 3D Cartesian space via translations and rotations~\cite{zhao2023representation, zitkovich2023rt, kim2024openvla, chi2025diffusion, goyal2023rvt, levine2016end, ze20243d, ze2025generalizable}, often using continuous 6D rotation parameterizations for numerical stability~\cite{zhou2019continuity, zech2019action}. Such representations are hard to interpret visually and decouple action structure from image observations. Point-cloud-based action representations~\cite{pan2023tax, huang2024imagination, huang2025match} are typically restricted to open-loop pick-and-place. Keypoint-based representations~\cite{choi2010real, huang2024rekep, patel2025real, liu2025kuda, tian2024robokeygen, ren2025motion} unify translation and rotation compactly, but when paired with 2D image inputs they generally place actions and observations in separate spaces. Image-based affordances~\cite{bahl2023affordances, montesano2009learning, manuelli2019kpam} and action-centric pixel representations~\cite{zeng2021transporter, huang2022equivariant, shridhar2022cliport, jia2025learning} ground only translation, leaving rotation ungrounded. Closest to ours, \citet{ren2025motion} project 3D actions onto fixed agent-view cameras and predict discretized pixel actions from single views, but suffer from discretization, triangulation errors under distant viewpoints, and cross-view inconsistencies. In contrast, we project 3D actions onto two image planes as continuous, unbounded coordinates, grounding 3D actions in 2D image space without precision loss.

\textbf{Symmetric manipulation learning.} Manipulation tasks are typically symmetric under translations, rotations, and reflections, and equivariant networks provide tools for encoding such symmetries~\cite{e2cnn, deng2021vector, cesa2022a, liao2022equiformer, he2021efficient}. Prior work has exploited symmetries for grasping~\cite{zhu2022grasp, zhu2023robot, huang2023edge, huorbitgrasp}, deformable and articulated manipulation~\cite{yang2024equivact}, efficient planning~\cite{zhao2022integrating, zhao2024mathrm}, and pick-and-place~\cite{wang2021q, simeonov2022neural, simeonov2023se, Huang-RSS-22, huang2024leveraging, huang2024fourier, ryu2022equivariant, ryu2023diffusion, pan2023tax, eisner2024deep}. More recent work realizes equivariant closed-loop policies with improved sample efficiency~\cite{wang2022so2equivariant, jia2023seil, wang2024equivariant, wang2022onrobot, liu2023continual, kohler2023symmetric, nguyen2023equivariant, nguyen2024symmetry, yang2024equivact, yang2024equibot}. In contrast, we formulate \textit{$n$-equivariance} within the policy learning framework and encourage it via data augmentation rather than equivariant architectural constraints.

\section{Problem Statement}
This work mainly focuses on imitation learning. 
Given image observations $O_t$ that capture the current environment using $n$ cameras $O_t=(o_{1_t}, o_{2_t}\cdots,o_{n_t})$,  we aim to learn a mapping $\pi \colon O_t \rightarrow A_t$ from observations to an action chunk  $A_t=(a_t,\cdots,a_{t+h})$ representing an action sequence with horizon $h$. Specifically, each single action is parameterized as $a=(T,R,w)$, where $T\in \mathbb{R}^3$, $R\in \mathrm{SO}(3)$ and $w\in \mathbb{R}$ denote the translation, rotation and gripper aperture, respectively. In addition to visual observations, the policy may also consume low-level proprioceptive inputs.
In the following sections, we first describe the process of grounding action chunks onto image planes as 2D image coordinates, and then formulate the policy learning framework based on this action representation.
\section{Method}
\subsection{Representing 3D actions as 2D keypoints}
\label{3d_2d_transformation}
\textbf{From Rotation to Translation.} 
Inspired by~\cite{ren2025motion,haldar2026point}, we first create a bijective mapping $\mathcal{M}$ from rigid motions in $\mathrm{SE}(3)$ described by homogeneous matrices to 3D keypoint representations $\mathcal{M}\colon(T,R) \mapsto (p^1,\cdots, p^m)$, where $p^i\in \mathbb{R}^3$ denotes a 3D keypoint on the gripper.  The choice of keypoints is flexible. Specifically, we select four keypoints along the gripper fingers to define the gripper pose
as illustrated in Figure~\ref{fig:action_repr}a.
It is worth noting that modeling rigid bodies using keypoints is not restricted to parallel-jaw grippers; extending this representation to other hardware configurations is possible but beyond the scope of this work.

Given the keypoints, the inverse mapping is defined as follows (Figure~\ref{fig:action_repr}a). The translation component $T$ can be calculated as
    $T = \frac{p^1+p^2+ p^3+ p^4}{4}$.
The rotation matrix $R$ can be constructed directly from the antipodal axis, $v_\mathrm{antipodal} = \frac{(p^2-p^1) + (p^4-p^3)}{2}$, and the approach axis, $v_\mathrm{approach} = \frac{(p^3-p^1) + (p^4-p^2)}{2}$, by conducting Gram–Schmidt orthonormalization according to the right-hand rule.
The gripper width is not reflected in the keypoint configuration and is instead represented as a scalar itself.
This transforms the action chunk from a sequence of $(R, T, w)$ into a sequence of $\big(p^1, p^2, p^3, p^4, w\big)$.

\textbf{From 3D keypoints to 2D keypoints.}
We then define another pair of maps
$(\mathcal{P},\mathcal{T})$ that transform between 3D points and 2D pixel coordinates using projection and triangulation: $\mathcal{P}[c_1, \cdots, c_n]\colon p^{j} \mapsto (\mathrm{pix}_1^j,\cdots, \mathrm{pix}^j_n)$ and $\mathcal{T}[c_1, \cdots, c_n]\colon (\mathrm{pix}_1^j,\cdots, \mathrm{pix}_n^j) \mapsto p^{j}$, where
$c_k$ denotes the $k$-th camera matrix and $\mathrm{pix}_k^j$ represents the projected image coordinate of the $j$-th 3D keypoint on the $k$-th camera plane.
Note that for generic camera positions $(\mathcal{P},\mathcal{T})$ are bijections $\mathbb{R}^3 \leftrightarrow \mathrm{Im}(\mathcal{P})$ and remain \emph{invertible} even under imperfect camera calibration.
Projection and triangulation are defined across the unbounded camera plane without clamping to the image boundaries. We then denote the pixel representation of the action chunk $A$ as \emph{image action chunk}:
\begin{equation}
    A_{\mathrm{pix}} = \{\mathrm{pix}_{ijk}\}_{i=1,j=1,k=1}^{h,m,n}
\end{equation}
where $i$, $j$, and $k$ denote the indices of the time horizon, keypoints, and cameras, respectively.

\subsection{Symmetry Properties and Equivariant Data augmentation}

The image action chunk $A_{\mathrm{pix}}$ transforms the policy function $\pi$ from a high-dimensional 3D control problem into a 2D trajectory prediction problem. Specifically,
given a robot system with $n$ cameras, the visuo-motor policy function $\pi$ can be  formulated as generating image coordinate actions on $n$ camera planes:
\begin{equation}
    \pi(o_1, o_2,\cdots,o_n) =  (A_{\mathrm{pix}}^1,A_{\mathrm{pix}}^2,\cdots, A_{\mathrm{pix}}^n)
    \label{policy_learning}
\end{equation}
where $ A_{\mathrm{pix}}^k$ denotes the action chunk on the k-th camera projection plane.
In the following, we first introduce the concept of \textit{$n$-equivariance} associated with Equation~\ref{policy_learning}, and then describe the corresponding equivariant data augmentation used to realize this property during the policy training. 

\textbf{The $n$-Equivariance Property.}
 There are two symmetry properties associated with policies defined with respect to image coordinate actions: (1) roto-translations of each camera in the camera plane and (2) permutation of the cameras. 

\begin{proposition}
\label{prop1} 
Given a list of individual $\SE(2)$ transformations $(g_1, g_2, \ldots, g_n)$ applied to the observation camera-by-camera, the desired action chunk produced by the policy should be transformed accordingly:
\begin{equation*}
\pi(g_1\cdot o_1,\cdots,g_n\cdot o_n) =  (g_1\cdot A_{\mathrm{pix}}^1,\cdots, g_n\cdot A_{\mathrm{pix}}^n)
\end{equation*}
\vspace{-0.4em}
\end{proposition}
\vspace{-0.4em}
Proposition~\ref{prop1} indicates that, since the observation and the action live in the same space, transformations applied to the observation are equivariantly also applied in the action space. Although the image observations are rotated, they convey the same semantic context and geometric relationships. An optimal policy should therefore be able to handle such transformations and learn invariant structures across them. Detailed analysis can be found in the Appendix~\ref{appendix-equivariance}.

\begin{proposition}
    If a permutation matrix $P \in \mathbb{R}^{n \times n}$ is applied to the observation, the generated action chunk undergoes the corresponding permutation.
    \begin{equation*}
\pi\left(\!\!P\cdot\begin{bmatrix}
o_1 & 0   & \cdots & 0 \\
0   & o_2    & \cdots & 0 \\
\vdots & \vdots & \ddots & \vdots \\
0   & 0      & \cdots & o_n
\end{bmatrix}\right) = P\cdot\pi\left(\begin{bmatrix}
o_1 & 0   & \cdots & 0 \\
0   & o_2    & \cdots & 0 \\
\vdots & \vdots & \ddots & \vdots \\
0   & 0      & \cdots & o_n
\end{bmatrix}\right)
    \end{equation*}
    \label{prop2}
\end{proposition}
Proposition~\ref{prop2} implies the optimal policy is permutation-equivariant: transforming one image equivariantly transforms its corresponding image action chunk while leaving the others unchanged. The two propositions together define the \textit{$n$-equivariance} underlying the policy learning $\pi(G\cdot O) = G\cdot\pi(O)$,
where $G \in \SE(2)^n \rtimes S_n$ is the combined group of $n$ independent planar rigid motions with the permutation group on $n$ elements. This group is known as a wreath product $\SE(2) \wr S_n$.
If we can train a policy to satisfy $n$-equivariance, the policy model learns from equivalence classes of demonstrations rather than isolated samples. This motivates architectures that explicitly respect the $\SE(2)^n \rtimes S_n$ symmetry, enabling consistent transformation of multi-view observations and image action chunks. Note that at the level of image-space action chunks, $n$-equivariance implies that the triangulated policy $\mathcal{T} \circ \pi$ is \emph{invariant} to $\SE(2)^n \rtimes S_n$ (acting jointly on observations and camera poses), since the transformations cancel out during triangulation. In other words, $n$-equivariance enforces invariance of the 3D actions to per-camera roto-translations and camera permutations.

\begin{figure*}[t]
    \centering
    \includegraphics[width=0.95\linewidth]{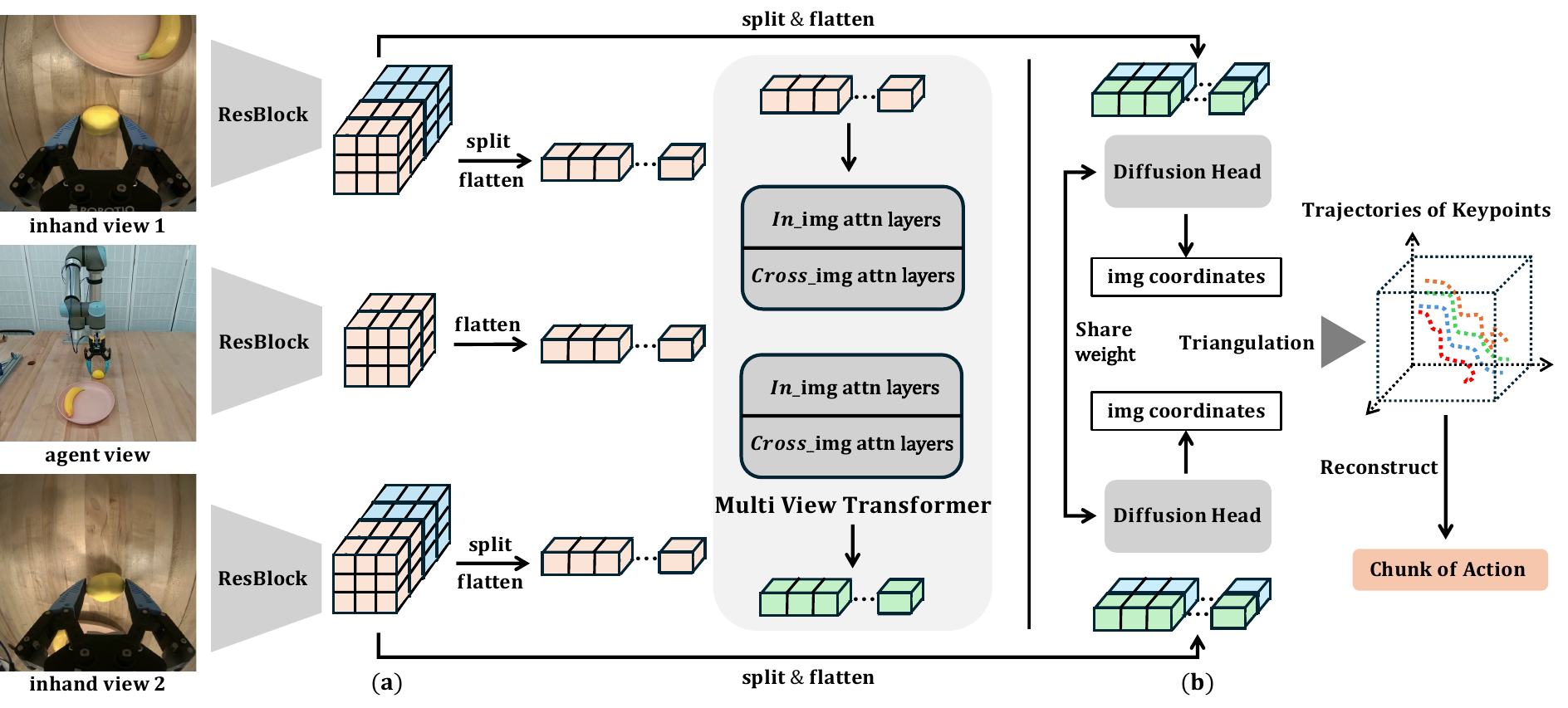}
    \caption{\textbf{Architecture of Diffusion X-Net.} \textbf{(a)} Observation Encoding; \textbf{(b)} Action Generation. The left branches of X-Net take two in-hand images and one side-view image as input. The center of X-Net features a multi-view transformer that enables communication between the in-hand features and the side-view context features. The right branches of X-Net regress two sets of continuous keypoints separately from the two in-hand features with $\SE(2)$ augmentation. As a result, our model learns an $n$-equivariant function.}
    \label{fig:archi}
\end{figure*}

Fortunately, $ A_{\mathrm{pix}}^k$ facilitates data augmentation that consistently transforms the observation and its corresponding image action chunk. By sampling $\SE(2)$ transformations, we enforce the policy model to satisfy Proposition~\ref{prop1}, effectively enlarging the support of the data distribution and allowing the model to learn the inherent group orbits across multi-image transformations. In the following section, we introduce our hardware system and architectural design that leverage Proposition~\ref{prop2}.

\subsection{Policy Learning on Image Action Chunk}
\textbf{Dual In-hand Camera Setup.}
To better align the image action representation with the robot embodiment, we mount two in-hand cameras on either side of the gripper, as shown in Figure~\ref{fig:action_repr}a, complemented by a head-mounted camera that provides global context. Representative image observations are shown on the left side of Figure~\ref{fig:archi}. 
We aim to learn a policy that consumes dual in-hand observations together with an agent-view image, and predicts the image action chunk on the in-hand cameras: $\pi(o_1, o_2, o_3) = (A_{\mathrm{pix}}^1,A_{\mathrm{pix}}^2)$.
We augment the two in-hand observations consistently with their corresponding image action chunks, $(o_1, A_{\mathrm{pix}}^1)$ and $(o_2, A_{\mathrm{pix}}^2)$, while transformations to the head-camera observation $o_3$ are applied independently. Augmenting the head camera independently makes the learning problem harder but encourages the policy to be \emph{invariant} to head-camera perturbations, while remaining \emph{equivariant} to rotations of the in-hand cameras\footnote{Transformations applied to the agent-view camera loosely mimic the viewpoint changes that may occur during deployment, encouraging the policy to learn invariance (robustness) to such motion.}.
Additionally, the bilateral symmetry of the parallel-jaw gripper induces a useful redundancy: rotating the gripper by $180^\circ$ around its approach axis simply swaps the two in-hand camera views. The permutation equivariance in Proposition~\ref{prop2} ensures that the recovered 3D action is invariant to this swap, factoring out this symmetry from the observation space\footnote{This property is specific to parallel-jaw grippers and does not account for low-level proprioceptive states or kinematic constraints.}.

\textbf{Architecture Design.}
Since policy learning with image action chunks inherits a multimodal distribution in a continuous action space, we model the policy as a diffusion process~\cite{ho2020denoising}, 
$ \epsilon_\theta(O, A_{\mathrm{pix}}^{\tau}, \tau)$, using a network parameterized by $\theta$.  During training, pairs $(O, A_{\mathrm{pix}})$ are sampled from the expert dataset, and scheduled Gaussian noise $\epsilon^{\tau}$ is added to $A_{\mathrm{pix}}$. The training objective is defined as $\mathcal{L}=||\epsilon_\theta(O, A_{\mathrm{pix}}+\epsilon^{\tau},\tau) - \epsilon^{\tau}||^2$. During inference, the policy performs a sequence of denoising steps, starting from a randomly sampled action $A_{\mathrm{pix}} \sim \mathcal{N}(0, I)$ and iteratively subtracting the predicted noise to generate a clean action. 

We adopt an X-shaped architecture, as shown in Figure~\ref{fig:archi}. First, the image observations are processed by a stack of ResBlocks that progressively trade spatial resolution for channel dimensionality. The encoded in-hand features are then split along the channel dimension and flattened into sequences of tokens. At the center of the X-Net, a multi-view transformer~\cite{goyal2023rvt,hamdi2021mvtn} processes one split from each in-hand feature stream together with context tokens derived from the agent-view observation. The multi-view Transformer consists of in-image attention layers and cross-image attention layers, which encourage information exchange across views. In addition, low-level proprioceptive information can be incorporated as tokens and fed into the cross-attention layers. 
The right branch of the X-Net is implemented as a shared, per-view diffusion Transformer that takes all tokens from each in-hand view and predicts the corresponding image action chunk. Notably, the per-view diffusion heads encourage independent observation-to-action prediction for each view while enforcing the permutation equivariance of Prop~\ref{prop2}. Full implementations are provided in Appendix~\ref{implementation-details} and~\ref{appendix-training-details}.

The policy is trained without access to 3D pose information, relying solely on image action chunks. Extensive equivariant data augmentation is applied to jointly augment observations and actions during training. During inference, the predicted image coordinates are triangulated using the known camera matrices to recover 3D keypoint trajectories. These 3D trajectories are then used to reconstruct the 3D action chunk (Section~\ref{3d_2d_transformation}), which can be sent directly to the robot at a fixed control frequency.
\begin{figure*}[t]
    \centering
         \centering
     \begin{subfigure}[b]{0.092\textwidth}
         \centering
         \includegraphics[width=0.99\textwidth]{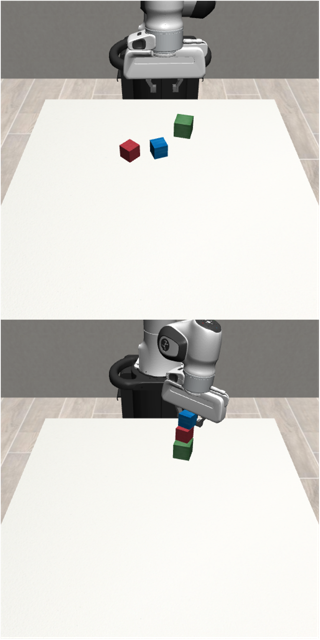}
     \end{subfigure}
     \begin{subfigure}[b]{0.092\textwidth}
         \centering
         \includegraphics[width=0.99\textwidth]{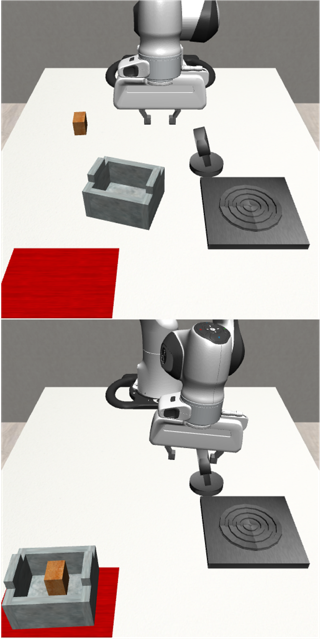}
     \end{subfigure}
     \begin{subfigure}[b]{0.092\textwidth}
         \centering
         \includegraphics[width=0.99\textwidth]{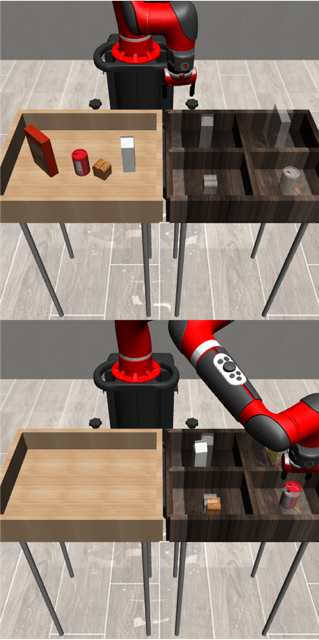}
     \end{subfigure}
     \begin{subfigure}[b]{0.092\textwidth}
         \centering
         \includegraphics[width=0.99\textwidth]{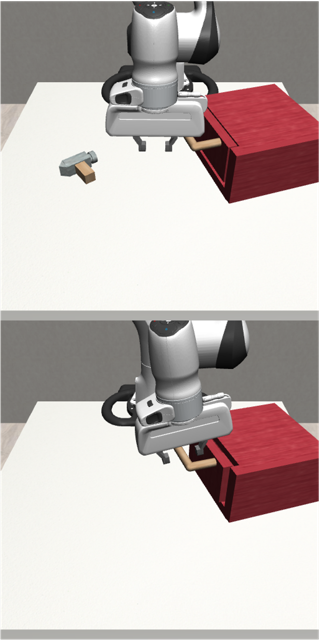}
     \end{subfigure}
      \begin{subfigure}[b]{0.092\textwidth}
         \centering
         \includegraphics[width=0.99\textwidth]{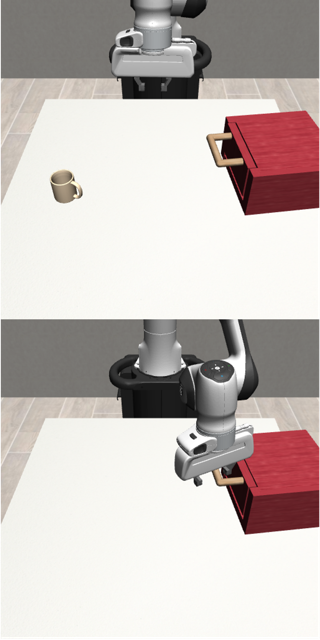}
     \end{subfigure}
      \begin{subfigure}[b]{0.092\textwidth}
         \centering
         \includegraphics[width=0.99\textwidth]{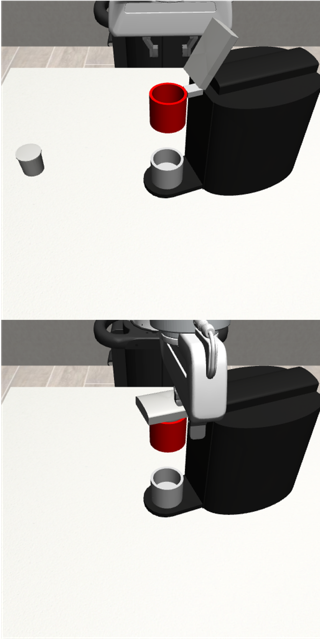}
     \end{subfigure}
      \begin{subfigure}[b]{0.092\textwidth}
         \centering
         \includegraphics[width=0.99\textwidth]{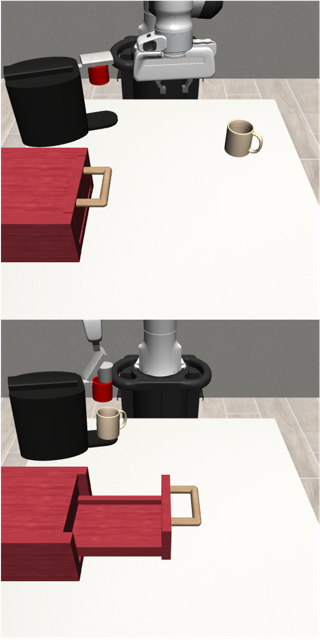}
     \end{subfigure}
      \begin{subfigure}[b]{0.092\textwidth}
         \centering
         \includegraphics[width=0.99\textwidth]{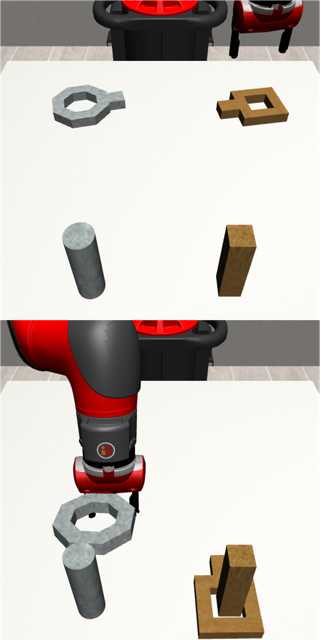}
     \end{subfigure}
      \begin{subfigure}[b]{0.092\textwidth}
         \centering
         \includegraphics[width=0.99\textwidth]{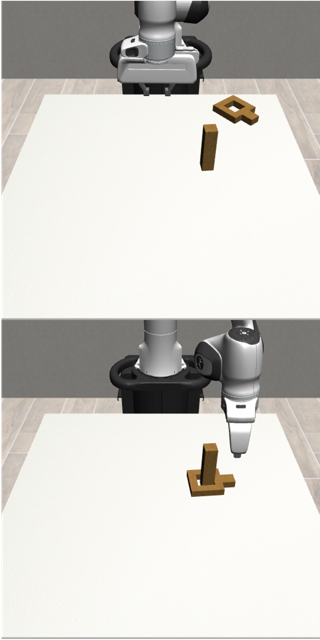}
     \end{subfigure}
      \begin{subfigure}[b]{0.092\textwidth}
         \centering
         \includegraphics[width=0.99\textwidth]{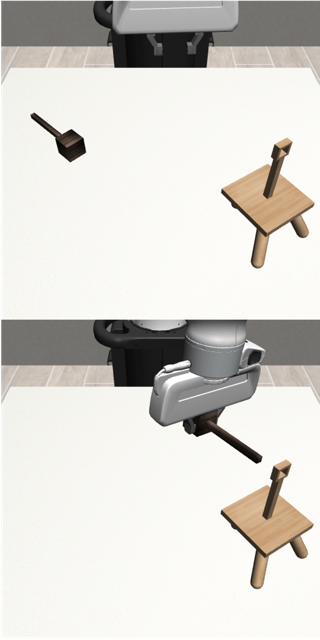}
     \end{subfigure}
     
    \caption{\textbf{Manipulation Tasks from Mimicgen.} The upper row and the bottom row show the initial states and the goal states.}
    \label{fig:mimicgen}
\end{figure*}
\begin{table*}[t]
  \setlength\tabcolsep{3pt}
  \begin{center}
  \scriptsize
  \begin{tabular}{@{}lcccccccccc c@{}}
  \toprule
  Method & StackThree & Kitchen & PickPlace & Hammer & Mug & Coffee & CoffeePr. & NutAsm. & Square & Threading & Avg. \\
  \midrule
  Equidiff (Img)~\cite{wang2024equivariant}    & 55 & 67 & 42 & 65 & 49 & 60 & 77 & 74 & 25 & 22 & \textcolor{darkblue}{53.6} \\
  Equidiff (Voxel)~\cite{wang2024equivariant}  & 75 & \textbf{85} & 58 & 70 & 53 & 65 & \textbf{80} & 67 & 39 & 39 & \textcolor{darkblue}{63.1} \\
  DP3 (PCD)~\cite{ze20243d}                    & 7  & 45 & 12 & 54 & 21 & 34 & 10 & 16 & 7  & 12 & \textcolor{darkblue}{21.8} \\
  ACT~\cite{zhao2023learning}                  & 6  & 37 & 7  & 38 & 23 & 19 & 32 & 42 & 6  & 10 & \textcolor{darkblue}{22.0} \\
  Motion Track~\cite{ren2025motion}            & 4  & 12 & 18 & 30 & 24 & 36 & 8  & 9  & 14 & 12 & \textcolor{darkblue}{16.7} \\
  Diffusion Policy~\cite{chi2023diffusion}     & 16 & 40 & 26 & 40 & 38 & 20 & 38 & 47 & 6  & 12 & \textcolor{darkblue}{28.3} \\
  $\ours$                                      & \textbf{94} & 76 & \textbf{88} & \textbf{81} & \textbf{79} & \textbf{62} & \textbf{80} & \textbf{90} & \textbf{50} & \textbf{52} & \textcolor{darkblue}{\textbf{75.2}} \\
  \bottomrule
  \end{tabular}
  \end{center}
  \vspace{-0.5em}
  \caption{\textbf{Performance Comparisons on the MimicGen Benchmark.} Success rate (mean \%) over 50 unseen tests, with 100 demonstration episodes used for training. Task suffixes (d0/d1/d2) follow the MimicGen variants described in Section~\ref{simulation-experiment}.}
  \label{table:single-task}
\end{table*}

\section{Experiment}
\subsection{Simulation Experiment}
\label{simulation-experiment}
\textbf{Task Description.} We evaluate $\ours$ on 10 MimicGen tasks~\cite{mandlekar2023mimicgen} (Figure~\ref{fig:mimicgen}), grouped into three categories:
(1) \textit{stack-three-d1}, \textit{kitchen-d1}, and \textit{pick-place-d0} primarily involve pick-place level precision;
(2) \textit{hammer-cleanup-d1}, \textit{mug-cleanup-d1}, \textit{coffee-d2}, and \textit{coffee-preparation-d1} require higher precision and involve manipulation of articulated objects;
(3) \textit{nut-assembly-d0}, \textit{square-d2}, and \textit{threading-d2} demand very high precision. 
The d0/d1/d2 suffixes denote light, moderate, and substantial spatial variation in the test distribution, and we use the largest variant provided by MimicGen~\cite{mandlekar2023mimicgen} for each task. See Appendix~\ref{appendix-sim} for full task descriptions.

\textbf{Setting and metrics.}
The system uses three cameras: two in-hand cameras mounted on the end-effector and one agent-view camera, all at $128 \times 128$ resolution. Our method predicts an action chunk of 12 steps and executes the first 8 at each rollout. All models are trained for 800 epochs on 100 demonstrations per task and evaluated every 20 epochs on 50 unseen test episodes; we report the best mean success rate across these checkpoints.

\textbf{Baseline.}
We compare our method against several representative baselines: (1) \textbf{Diffusion Policy}~\cite{chi2023diffusion}, the standard $\SE(3)$ action-chunk diffusion baseline using a 6D rotation parameterization; (2) \textbf{EquiDiff}~\cite{wang2024equivariant}, an equivariant variant that exploits rotational symmetries in $\SE(3)$ action space, evaluated in both image and voxel forms; (3) \textbf{ACT}~\cite{zhao2023learning}, a Transformer-based policy that encodes $\SE(3)$ action chunks as style vectors via a conditional VAE; (4) \textbf{DP3}~\cite{ze20243d}, a 3D diffusion policy operating on point clouds, which provides direct access to 3D geometry beyond what 2D images offer; and (5) \textbf{Motion Track}~\cite{ren2025motion}, a representative image-keypoint diffusion method that denoises pixel coordinates per view and triangulates across views to recover 3D actions. We use the authors' original implementations. Results for Equidiff~\cite{wang2024equivariant}, ACT~\cite{zhao2023learning}, and DP3~\cite{ze20243d} are taken from~\cite{wang2024equivariant}. Please note  we focus on single-task policy performance, and no method uses a pretrained vision encoder.
\begin{wrapfigure}[9]{r}{0.70\textwidth}
  \vspace{-1em}
  \centering
  \includegraphics[width=0.70\textwidth]{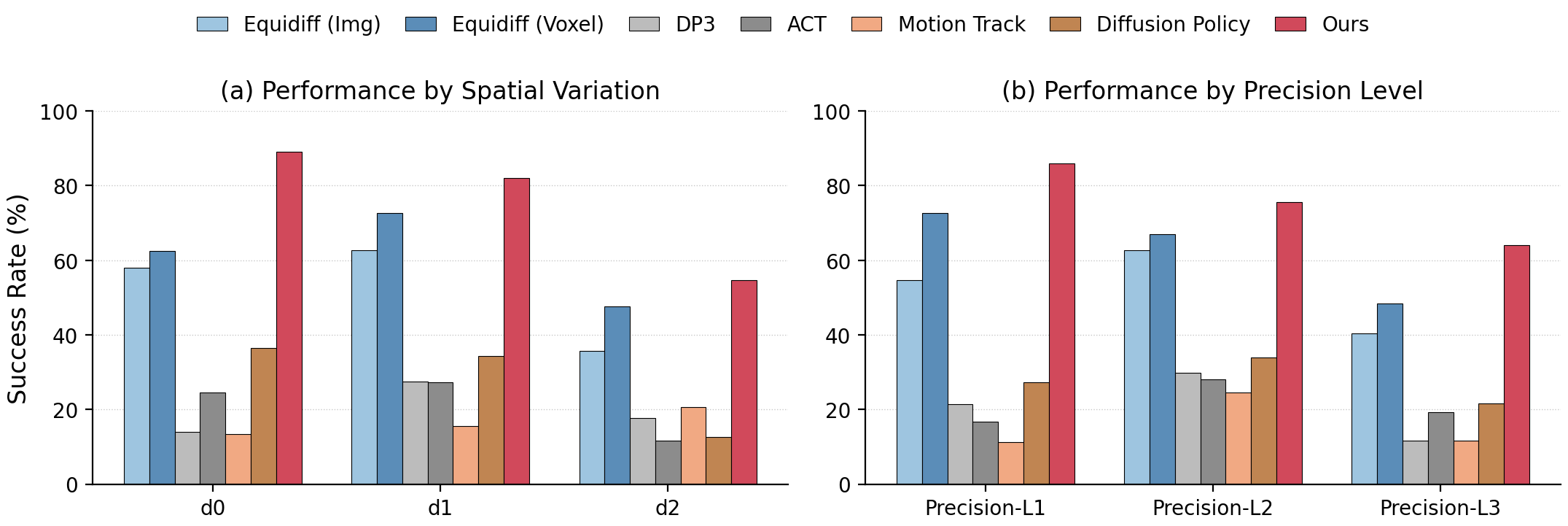}
  \caption{{Performance Comparisons on Different Types of Task.}}
  \label{fig:grouped-comparison}
\end{wrapfigure}
\textbf{Results.}
Table~\ref{table:single-task} summarizes the results. \textbf{First}, $\ours$ leads on 9/10 tasks with 75.2\% average success (100 demos), 12.1\% above the best baseline (EquiDiff-Voxel at 63.1\%), 21.6\% above the 
best image baseline (EquiDiff-Img at 53.6\%). This highlights the effectiveness of our image-based action representation, which provides a compact, task-aligned parameterization. \textbf{Second}, as task precision and spatial variation requirements increase, $\ours$ retains the largest gains over all baselines (Figure~\ref{fig:grouped-comparison}), supporting our analysis that $n$-equivariance, enforced through both the action representation and Diffusion X-Net, yields more robust generalization under spatial perturbations. \textbf{Third}, $\ours$ significantly outperforms Motion Track, whose independent per-view inference produces inconsistent trajectories that our cross-view design avoids. We additionally provide a multi-task extension in Appendix~\ref{multi-task-extension}.
\begin{wraptable}{r}{0.55\textwidth}
  \vspace{-0.4em}
  \setlength\tabcolsep{2.5pt}
  \scriptsize
  \begin{tabular}{@{}lccccc@{}}
  \toprule
  Method & StackThree & Hammer & Square & Threading & Avg. \\
  \midrule
  $\ours$                  & \textbf{94} & 81 & \textbf{50} & \textbf{52} & \textbf{69.3} \\
  \,\, w/o equiv. aug.     & 76\,\textsubscript{($\downarrow$18)} & 62\,\textsubscript{($\downarrow$19)} & 30\,\textsubscript{($\downarrow$20)} & 36\,\textsubscript{($\downarrow$16)} & 51.0\,\textsubscript{($\downarrow$18.3)} \\
  \,\, w/o shared head     & 92\,\textsubscript{($\downarrow$2)}  & 80\,\textsubscript{($\downarrow$1)}  & 8\,\textsubscript{($\downarrow$42)}  & 18\,\textsubscript{($\downarrow$34)} & 49.5\,\textsubscript{($\downarrow$19.8)} \\
  \,\, 3D action (abs.)    & 50\,\textsubscript{($\downarrow$44)} & 70\,\textsubscript{($\downarrow$11)} & 18\,\textsubscript{($\downarrow$32)} & 28\,\textsubscript{($\downarrow$24)} & 41.5\,\textsubscript{($\downarrow$27.8)} \\
  \,\, 3D action (rel.)    & 88\,\textsubscript{($\downarrow$6)}  & 80\,\textsubscript{($\downarrow$1)}  & 30\,\textsubscript{($\downarrow$20)} & 24\,\textsubscript{($\downarrow$28)} & 55.5\,\textsubscript{($\downarrow$13.8)} \\
  \,\, w/o agent view      & 90\,\textsubscript{($\downarrow$4)}  & 78\,\textsubscript{($\downarrow$3)}  & 50\,\textsubscript{($-$)}            & 48\,\textsubscript{($\downarrow$4)} & 66.5\,\textsubscript{($\downarrow$2.8)} \\
  \,\, w/o proprio         & 90\,\textsubscript{($\downarrow$4)}  & \textbf{82}\,\textsubscript{($\uparrow$1)} & 48\,\textsubscript{($\downarrow$2)}  & 52\,\textsubscript{($-$)}            & 68.0\,\textsubscript{($\downarrow$1.3)} \\
  \bottomrule
  \end{tabular}
  \caption{\textbf{Ablation study.} Success rate (mean \%) on 50 unseen tests, 100 demonstrations.}
  \label{table:ablation}
  \vspace{-1em}
\end{wraptable}
\textbf{Ablation Study.}
We analyze the importance of our design choices through a series of ablations.
(1) We disable equivariant data augmentation to isolate its contribution.
(2) We replace the shared per-camera diffusion heads with a single joint head, testing the value of permutation equivariance.
(3) We bypass the image action representation altogether, feeding all X-Net tokens directly into the diffusion head to predict 3D action chunks; we consider two variants: absolute actions, and relative actions canonicalized by the end-effector pose.
(4) We remove the head-mounted (agent-view) camera to assess the role of the third view.
(5) We compare $\ours$ with and without low-level proprioceptive state input to quantify its impact.
We report the ablation results in Table~\ref{table:ablation}. \textbf{First}, removing equivariant data augmentation causes an 18.3\% average drop, confirming its role in forcing the model to learn geometry-consistent features across views. \textbf{Second}, replacing the shared per-camera diffusion head with a joint head yields a 19.8\% average drop, with larger degradation on harder tasks (Square, Threading), supporting that permutation equivariance helps the model better exploit per-view augmentations. \textbf{Third}, directly predicting 3D action chunks from X-Net tokens, in either absolute or relative form, degrades performance by 27.8\% and 13.8\% respectively, validating the benefit of grounding actions in image space rather than $\SE(3)$. \textbf{Fourth}, removing the agent view has a slight effect, suggesting it primarily provides complementary context. \textbf{Finally}, $\ours$ remains nearly unchanged without proprioceptive state input (1.3\% drop).

\subsection{Real-world Experiment}

\begin{figure*}[t]
  \centering
  \begin{subfigure}[t]{0.245\textwidth}
    \includegraphics[width=\linewidth]{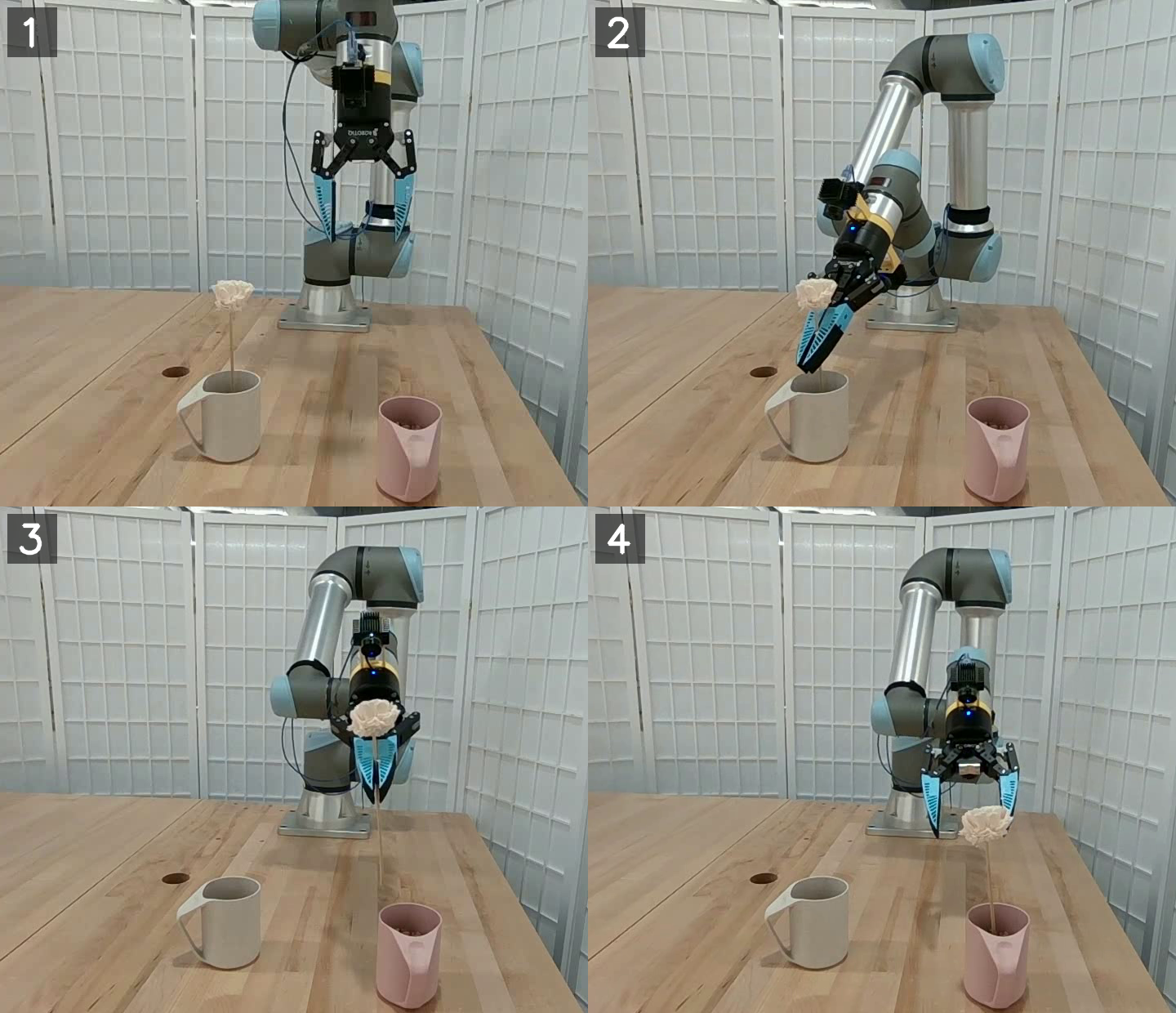}
    \caption{Plug Flower}
    \label{fig:real-a}
  \end{subfigure}\hfill
  \begin{subfigure}[t]{0.245\textwidth}
    \includegraphics[width=\linewidth]{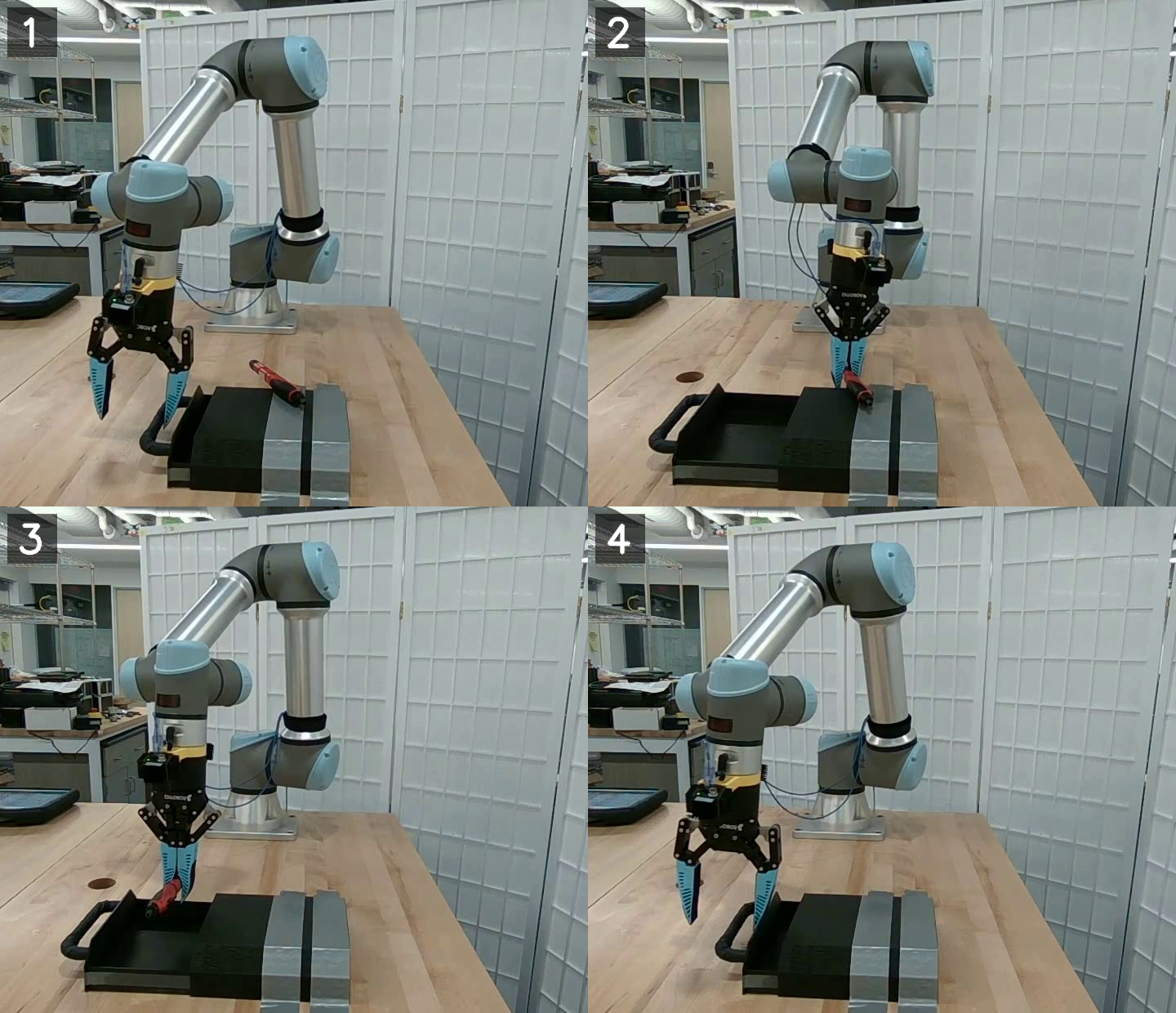}
    \caption{Put in Drawer}
    \label{fig:real-b}
  \end{subfigure}\hfill
  \begin{subfigure}[t]{0.245\textwidth}
    \includegraphics[width=\linewidth]{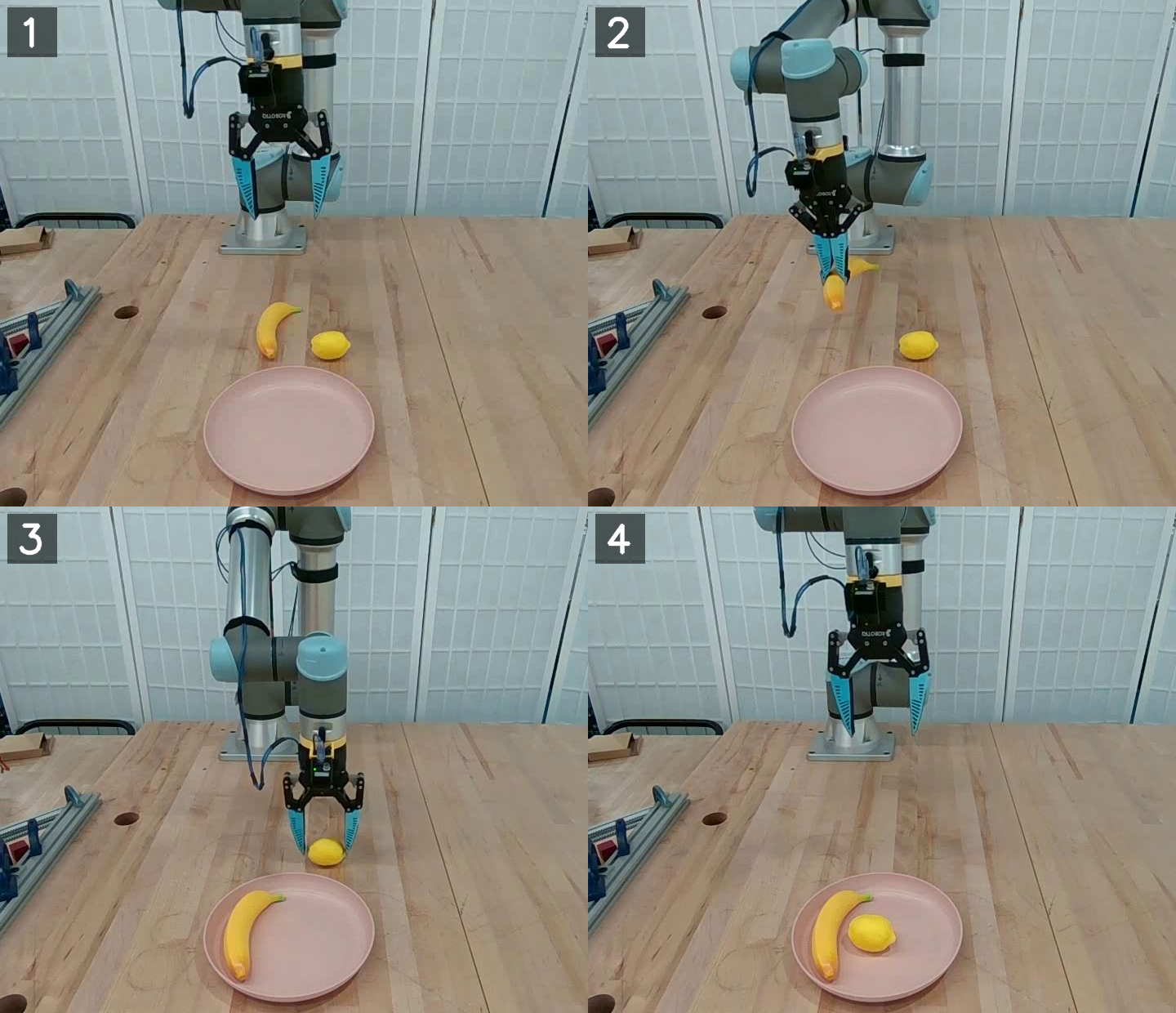}
    \caption{Prepare Fruit Plate}
    \label{fig:real-c}
  \end{subfigure}\hfill
  \begin{subfigure}[t]{0.245\textwidth}
    \includegraphics[width=\linewidth]{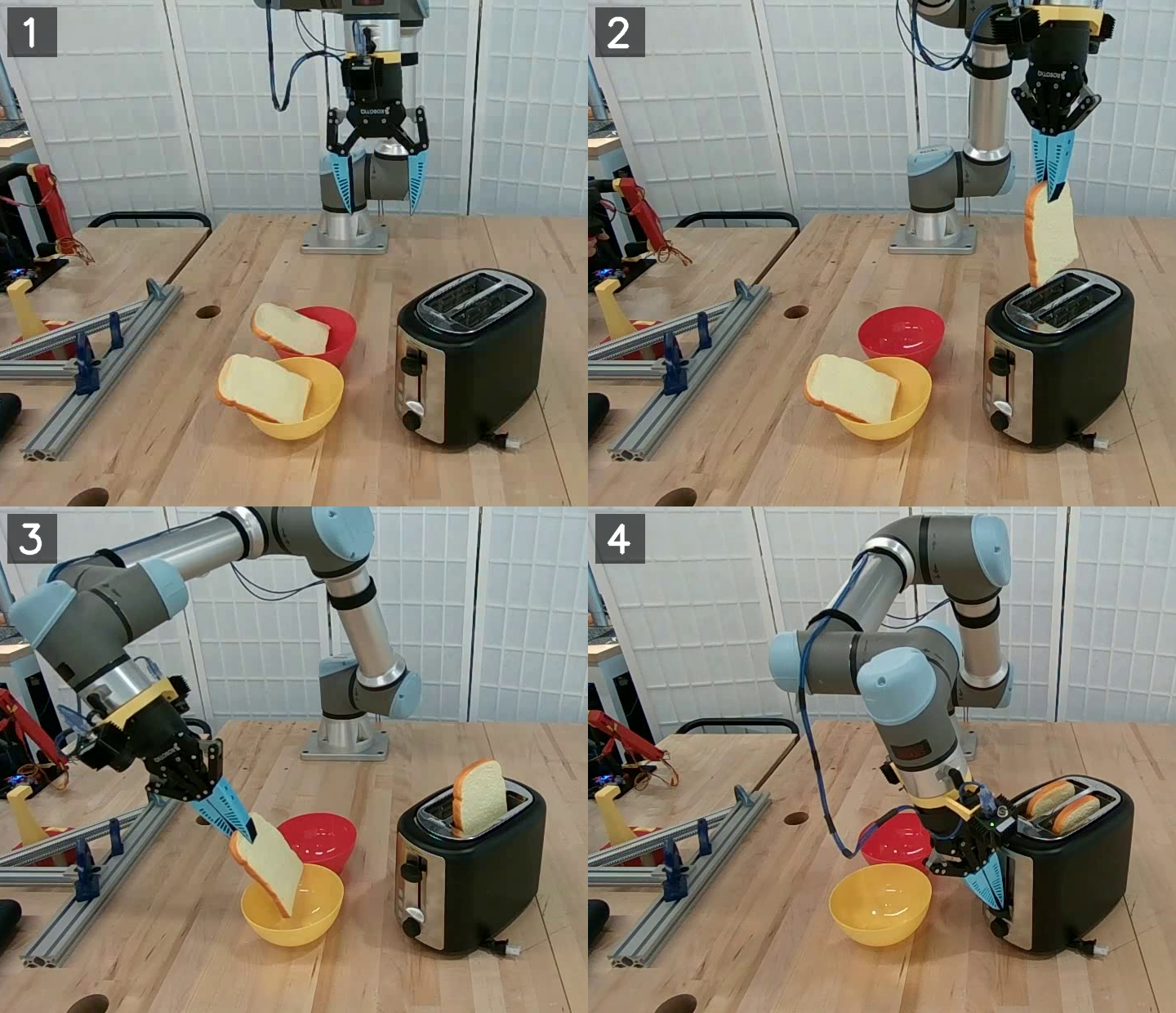}
    \caption{Make Toast}
    \label{fig:real-d}
  \end{subfigure}
  \caption{\textbf{Real-world manipulation tasks.} The four single-arm tasks evaluated on our UR5 platform.}
  \label{fig:real-world}
\end{figure*}

We conduct real-world experiments on four tasks, focusing on single-task performance and training all methods from scratch. As shown in Figure~\ref{fig:real-world}, we use a UR5 robotic arm equipped with two in-hand cameras and one agent-view camera. \textbf{Plug Flower} requires the robot to grasp a flower and insert it into a base. \textbf{Put in Drawer} is a multi-step task involving opening a drawer, picking up a pen, placing it inside, and closing the drawer. \textbf{Prepare Fruit Plate} requires the robot to pick a banana and a lemon and place them on a plate. \textbf{Make Toast} requires the robot to grasp two slices of bread, insert them into the toaster, and press the button. We also measure the action reconstruction error of $(\mathcal{P}, \mathcal{T})$ using demonstration data, which is around $10^{-6}$ in our real-world setting.

\begin{wraptable}{r}{0.35\textwidth}
  \vspace{-0.4em}
  \setlength\tabcolsep{4pt}
  \scriptsize
  \centering
  \begin{tabular}{@{}lccc@{}}
    \toprule
    Task & \#Demo & $\ours$ & DP \\
    \midrule
    Plug Flower         & 30 & \textbf{80} & 55 \\
    Put in Drawer       & 60 & \textbf{90} & 50 \\
    Prepare Fruit Plate & 70 & \textbf{85} & 60 \\
    Make Toast          & 100 & \textbf{85} & 60 \\
    \bottomrule
  \end{tabular}
  \captionsetup{width=0.98\linewidth}
  \caption{\textbf{Real-world success rates} (\%) over 20 trials per task.}
  \label{table:real-world}
  \vspace{-1em}
\end{wraptable}
We compare $\ours$ against \textbf{Diffusion Policy}~\cite{chi2023diffusion}, with both methods trained using DDIM~\cite{song2020denoising} for faster inference. Table~\ref{table:real-world} reports completion rates over 20 trials per task with no partial credit. $\ours$ significantly outperforms the baseline across all tasks, consistent with our simulation results. We further evaluate robustness to camera perturbations on the \textbf{Plug Flower} and \textbf{Put in Drawer} tasks by randomly rotating the agent-view camera at test time. This differs from the agent-view ablation: rather than removing the view, we keep it but introduce out-of-distribution viewpoint shifts (Figure~\ref{fig:real-shake} in Appendix~\ref{appendix-realworld}). $\ours$ remains unaffected, while Diffusion Policy fails to complete either task, demonstrating the advantage of our image-space action representation and augmentation technique. More experiments with reduced demonstrations are provided in Appendix~\ref{appendix-realworld}.

\section{Conclusion and Limitation}
\label{sec:conclusion}

In this work, we propose $\ours$, an imitation learning framework for closed-loop robotic manipulation. Its key idea is a continuous, unbounded 2D image-based action representation, called the \emph{image action chunk}, which reformulates 3D policy learning as a 2D image-keypoint prediction problem. We analyze the symmetry properties of this representation and introduce the Diffusion X-Net architecture to align the model design with its structure. $\ours$ outperforms several strong baselines on 10 challenging MimicGen tasks and transfers effectively to a physical robot platform. Several limitations remain. First, it requires a calibrated dual-camera setup; under our design this is lightweight, assuming only a fixed offset between the cameras and the end-effector. Second, projected image coordinates can become discontinuous when the motion crosses the camera plane. We handle this carefully in our implementation (Appendix~\ref{implementation-details}) but leave more principled strategies to future work. Third, we focus on parallel-jaw grippers; extending the representation to dexterous hands is possible but beyond the scope of this paper. Finally, we train policies from scratch, and combining $\ours$ with a stronger pretrained vision encoder is a natural direction for future work.

\clearpage
\acknowledgments{This work was supported in part by the National Science Foundation under grants NSF 2442658, NSF 2314182, NSF 2107256 and NASA 80NSSC19K1474.}


\bibliography{example}  

@article{yan2025maniflow,
  title={Maniflow: A general robot manipulation policy via consistency flow training},
  author={Yan, Ge and Zhu, Jiyue and Deng, Yuquan and Yang, Shiqi and Qiu, Ri-Zhao and Cheng, Xuxin and Memmel, Marius and Krishna, Ranjay and Goyal, Ankit and Wang, Xiaolong and others},
  journal={arXiv preprint arXiv:2509.01819},
  year={2025}
}

@article{barreiros2025careful,
  title={A careful examination of large behavior models for multitask dexterous manipulation},
  author={Barreiros, Jose and Beaulieu, Andrew and Bhat, Aditya and Cory, Rick and Cousineau, Eric and Dai, Hongkai and Fang, Ching-Hsin and Hashimoto, Kunimatsu and Irshad, Muhammad Zubair and Itkina, Masha and others},
  journal={arXiv preprint arXiv:2507.05331},
  year={2025}
}

@article{song2019generative,
  title={Generative modeling by estimating gradients of the data distribution},
  author={Song, Yang and Ermon, Stefano},
  journal={Advances in neural information processing systems},
  volume={32},
  year={2019}
}

@article{alain2014regularized,
  title={What regularized auto-encoders learn from the data-generating distribution},
  author={Alain, Guillaume and Bengio, Yoshua},
  journal={The Journal of Machine Learning Research},
  volume={15},
  number={1},
  pages={3563--3593},
  year={2014},
  publisher={JMLR. org}
}

@article{ho2020denoising,
  title={Denoising diffusion probabilistic models},
  author={Ho, Jonathan and Jain, Ajay and Abbeel, Pieter},
  journal={Advances in neural information processing systems},
  volume={33},
  pages={6840--6851},
  year={2020}
}

@article{lipman2022flow,
  title={Flow matching for generative modeling},
  author={Lipman, Yaron and Chen, Ricky TQ and Ben-Hamu, Heli and Nickel, Maximilian and Le, Matt},
  journal={arXiv preprint arXiv:2210.02747},
  year={2022}
}

@article{mandlekar2023mimicgen,
  title={Mimicgen: A data generation system for scalable robot learning using human demonstrations},
  author={Mandlekar, Ajay and Nasiriany, Soroush and Wen, Bowen and Akinola, Iretiayo and Narang, Yashraj and Fan, Linxi and Zhu, Yuke and Fox, Dieter},
  journal={arXiv preprint arXiv:2310.17596},
  year={2023}
}

@article{song2020score,
  title={Score-based generative modeling through stochastic differential equations},
  author={Song, Yang and Sohl-Dickstein, Jascha and Kingma, Diederik P and Kumar, Abhishek and Ermon, Stefano and Poole, Ben},
  journal={arXiv preprint arXiv:2011.13456},
  year={2020}
}

@article{zhao2023learning,
  title={Learning fine-grained bimanual manipulation with low-cost hardware},
  author={Zhao, Tony Z and Kumar, Vikash and Levine, Sergey and Finn, Chelsea},
  journal={arXiv preprint arXiv:2304.13705},
  year={2023}
}

@inproceedings{
  wang2025practical,
  title={A Practical Guide for Incorporating Symmetry in Diffusion Policy},
  author={Dian Wang and Boce Hu and Shuran Song and Robin Walters and Robert Platt},
  booktitle={The Thirty-ninth Annual Conference on Neural Information Processing Systems},
  year={2025},
  url={https://openreview.net/forum?id=e0Dn7dg395}
}

@article{huang2022equivariant,
  title={Equivariant transporter network},
  author={Huang, Haojie and Wang, Dian and Walters, Robin and Platt, Robert},
  journal={arXiv preprint arXiv:2202.09400},
  year={2022}
}

@inproceedings{zeng2021transporter,
  title={Transporter networks: Rearranging the visual world for robotic manipulation},
  author={Zeng, Andy and Florence, Pete and Tompson, Jonathan and Welker, Stefan and Chien, Jonathan and Attarian, Maria and Armstrong, Travis and Krasin, Ivan and Duong, Dan and Sindhwani, Vikas and others},
  booktitle={Conference on Robot Learning},
  pages={726--747},
  year={2021},
  organization={PMLR}
}

@inproceedings{shridhar2022cliport,
  title={Cliport: What and where pathways for robotic manipulation},
  author={Shridhar, Mohit and Manuelli, Lucas and Fox, Dieter},
  booktitle={Conference on robot learning},
  pages={894--906},
  year={2022},
  organization={PMLR}
}

@article{jia2025learning,
  title={Learning efficient and robust language-conditioned manipulation using textual-visual relevancy and equivariant language mapping},
  author={Jia, Mingxi and Huang, Haojie and Zhang, Zhewen and Wang, Chenghao and Zhao, Linfeng and Wang, Dian and Liu, Jason Xinyu and Walters, Robin and Platt, Robert and Tellex, Stefanie},
  journal={IEEE Robotics and Automation Letters},
  year={2025},
  publisher={IEEE}
}

@article{fang2023anygrasp,
  title={Anygrasp: Robust and efficient grasp perception in spatial and temporal domains},
  author={Fang, Hao-Shu and Wang, Chenxi and Fang, Hongjie and Gou, Minghao and Liu, Jirong and Yan, Hengxu and Liu, Wenhai and Xie, Yichen and Lu, Cewu},
  journal={IEEE Transactions on Robotics},
  volume={39},
  number={5},
  pages={3929--3945},
  year={2023},
  publisher={IEEE}
}

@article{laskin2020reinforcement,
  title={Reinforcement learning with augmented data},
  author={Laskin, Misha and Lee, Kimin and Stooke, Adam and Pinto, Lerrel and Abbeel, Pieter and Srinivas, Aravind},
  journal={Advances in neural information processing systems},
  volume={33},
  pages={19884--19895},
  year={2020}
}

@article{hansen2020self,
  title={Self-supervised policy adaptation during deployment},
  author={Hansen, Nicklas and Jangir, Rishabh and Sun, Yu and Aleny{\`a}, Guillem and Abbeel, Pieter and Efros, Alexei A and Pinto, Lerrel and Wang, Xiaolong},
  journal={arXiv preprint arXiv:2007.04309},
  year={2020}
}

@inproceedings{shridhar2023perceiver,
  title={Perceiver-actor: A multi-task transformer for robotic manipulation},
  author={Shridhar, Mohit and Manuelli, Lucas and Fox, Dieter},
  booktitle={Conference on Robot Learning},
  pages={785--799},
  year={2023},
  organization={PMLR}
}

@article{goyal2024rvt,
  title={Rvt-2: Learning precise manipulation from few demonstrations},
  author={Goyal, Ankit and Blukis, Valts and Xu, Jie and Guo, Yijie and Chao, Yu-Wei and Fox, Dieter},
  journal={arXiv preprint arXiv:2406.08545},
  year={2024}
}

@article{wang2020policy,
  title={Policy learning in se (3) action spaces},
  author={Wang, Dian and Kohler, Colin and Platt, Robert},
  journal={arXiv preprint arXiv:2010.02798},
  year={2020}
}

@article{jia2022seil,
  title={Seil: Simulation-augmented equivariant imitation learning},
  author={Jia, Mingxi and Wang, Dian and Su, Guanang and Klee, David and Zhu, Xupeng and Walters, Robin and Platt, Robert},
  journal={arXiv preprint arXiv:2211.00194},
  year={2022}
}

@article{hu20253d,
  title={3D Equivariant Visuomotor Policy Learning via Spherical Projection},
  author={Hu, Boce and Wang, Dian and Klee, David and Tian, Heng and Zhu, Xupeng and Huang, Haojie and Platt, Robert and Walters, Robin},
  journal={arXiv preprint arXiv:2505.16969},
  year={2025}
}

@article{jiang2021synergies,
  title={Synergies between affordance and geometry: 6-dof grasp detection via implicit representations},
  author={Jiang, Zhenyu and Zhu, Yifeng and Svetlik, Maxwell and Fang, Kuan and Zhu, Yuke},
  journal={arXiv preprint arXiv:2104.01542},
  year={2021}
}

@inproceedings{tobin2017domain,
  title={Domain randomization for transferring deep neural networks from simulation to the real world},
  author={Tobin, Josh and Fong, Rachel and Ray, Alex and Schneider, Jonas and Zaremba, Wojciech and Abbeel, Pieter},
  booktitle={2017 IEEE/RSJ international conference on intelligent robots and systems (IROS)},
  pages={23--30},
  year={2017},
  organization={IEEE}
}

@inproceedings{robomimic2021,
  title={What Matters in Learning from Offline Human Demonstrations for Robot Manipulation},
  author={Ajay Mandlekar and Danfei Xu and Josiah Wong and Soroush Nasiriany and Chen Wang and Rohun Kulkarni and Li Fei-Fei and Silvio Savarese and Yuke Zhu and Roberto Mart\'{i}n-Mart\'{i}n},
  booktitle={Conference on Robot Learning (CoRL)},
  year={2021}
}

@article{dasari2019robonet,
  title={Robonet: Large-scale multi-robot learning},
  author={Dasari, Sudeep and Ebert, Frederik and Tian, Stephen and Nair, Suraj and Bucher, Bernadette and Schmeckpeper, Karl and Singh, Siddharth and Levine, Sergey and Finn, Chelsea},
  journal={arXiv preprint arXiv:1910.11215},
  year={2019}
}

@inproceedings{sennrich2016improving,
  title={Improving neural machine translation models with monolingual data},
  author={Sennrich, Rico and Haddow, Barry and Birch, Alexandra},
  booktitle={Proceedings of the 54th annual meeting of the association for computational linguistics (volume 1: long papers)},
  pages={86--96},
  year={2016}
}

@article{wei2021finetuned,
  title={Finetuned language models are zero-shot learners},
  author={Wei, Jason and Bosma, Maarten and Zhao, Vincent Y and Guu, Kelvin and Yu, Adams Wei and Lester, Brian and Du, Nan and Dai, Andrew M and Le, Quoc V},
  journal={arXiv preprint arXiv:2109.01652},
  year={2021}
}

@inproceedings{zhou2019continuity,
  title={On the continuity of rotation representations in neural networks},
  author={Zhou, Yi and Barnes, Connelly and Lu, Jingwan and Yang, Jimei and Li, Hao},
  booktitle={Proceedings of the IEEE/CVF conference on computer vision and pattern recognition},
  pages={5745--5753},
  year={2019}
}

@inproceedings{zitkovich2023rt,
  title={Rt-2: Vision-language-action models transfer web knowledge to robotic control},
  author={Zitkovich, Brianna and Yu, Tianhe and Xu, Sichun and Xu, Peng and Xiao, Ted and Xia, Fei and Wu, Jialin and Wohlhart, Paul and Welker, Stefan and Wahid, Ayzaan and others},
  booktitle={Conference on Robot Learning},
  pages={2165--2183},
  year={2023},
  organization={PMLR}
}

@article{kim2024openvla,
  title={Openvla: An open-source vision-language-action model},
  author={Kim, Moo Jin and Pertsch, Karl and Karamcheti, Siddharth and Xiao, Ted and Balakrishna, Ashwin and Nair, Suraj and Rafailov, Rafael and Foster, Ethan and Lam, Grace and Sanketi, Pannag and others},
  journal={arXiv preprint arXiv:2406.09246},
  year={2024}
}

@article{chi2025diffusion,
  title={Diffusion policy: Visuomotor policy learning via action diffusion},
  author={Chi, Cheng and Xu, Zhenjia and Feng, Siyuan and Cousineau, Eric and Du, Yilun and Burchfiel, Benjamin and Tedrake, Russ and Song, Shuran},
  journal={The International Journal of Robotics Research},
  volume={44},
  number={10-11},
  pages={1684--1704},
  year={2025},
  publisher={Sage Publications Sage UK: London, England}
}

@inproceedings{goyal2023rvt,
  title={Rvt: Robotic view transformer for 3d object manipulation},
  author={Goyal, Ankit and Xu, Jie and Guo, Yijie and Blukis, Valts and Chao, Yu-Wei and Fox, Dieter},
  booktitle={Conference on Robot Learning},
  pages={694--710},
  year={2023},
  organization={PMLR}
}

@article{levine2016end,
  title={End-to-end training of deep visuomotor policies},
  author={Levine, Sergey and Finn, Chelsea and Darrell, Trevor and Abbeel, Pieter},
  journal={Journal of Machine Learning Research},
  volume={17},
  number={39},
  pages={1--40},
  year={2016}
}

@article{zhao2023representation,
  title={Representation learning for continuous action spaces is beneficial for efficient policy learning},
  author={Zhao, Tingting and Wang, Ying and Sun, Wei and Chen, Yarui and Niu, Gang and Sugiyama, Masashi},
  journal={Neural Networks},
  volume={159},
  pages={137--152},
  year={2023},
  publisher={Elsevier}
}

@inproceedings{ze2025generalizable,
  title={Generalizable humanoid manipulation with 3d diffusion policies},
  author={Ze, Yanjie and Chen, Zixuan and Wang, Wenhao and Chen, Tianyi and He, Xialin and Yuan, Ying and Peng, Xue Bin and Wu, Jiajun},
  booktitle={2025 IEEE/RSJ International Conference on Intelligent Robots and Systems (IROS)},
  pages={2873--2880},
  year={2025},
  organization={IEEE}
}

@article{zech2019action,
  title={Action representations in robotics: A taxonomy and systematic classification},
  author={Zech, Philipp and Renaudo, Erwan and Haller, Simon and Zhang, Xiang and Piater, Justus},
  journal={The International Journal of Robotics Research},
  volume={38},
  number={5},
  pages={518--562},
  year={2019},
  publisher={SAGE Publications Sage UK: London, England}
}

@inproceedings{bahl2023affordances,
  title={Affordances from human videos as a versatile representation for robotics},
  author={Bahl, Shikhar and Mendonca, Russell and Chen, Lili and Jain, Unnat and Pathak, Deepak},
  booktitle={Proceedings of the IEEE/CVF Conference on Computer Vision and Pattern Recognition},
  pages={13778--13790},
  year={2023}
}

@inproceedings{montesano2009learning,
  title={Learning grasping affordances from local visual descriptors},
  author={Montesano, Luis and Lopes, Manuel},
  booktitle={2009 IEEE 8th international conference on development and learning},
  pages={1--6},
  year={2009},
  organization={IEEE}
}

@inproceedings{manuelli2019kpam,
  title={kpam: Keypoint affordances for category-level robotic manipulation},
  author={Manuelli, Lucas and Gao, Wei and Florence, Peter and Tedrake, Russ},
  booktitle={The International Symposium of Robotics Research},
  pages={132--157},
  year={2019},
  organization={Springer}
}

@inproceedings{choi2010real,
  title={Real-time 3D model-based tracking using edge and keypoint features for robotic manipulation},
  author={Choi, Changhyun and Christensen, Henrik I},
  booktitle={2010 IEEE international conference on robotics and automation},
  pages={4048--4055},
  year={2010},
  organization={IEEE}
}

@article{huang2024rekep,
  title={Rekep: Spatio-temporal reasoning of relational keypoint constraints for robotic manipulation},
  author={Huang, Wenlong and Wang, Chen and Li, Yunzhu and Zhang, Ruohan and Fei-Fei, Li},
  journal={arXiv preprint arXiv:2409.01652},
  year={2024}
}

@article{patel2025real,
  title={A real-to-sim-to-real approach to robotic manipulation with VLM-generated iterative keypoint rewards},
  author={Patel, Shivansh and Yin, Xinchen and Huang, Wenlong and Garg, Shubham and Nayyeri, Hooshang and Fei-Fei, Li and Lazebnik, Svetlana and Li, Yunzhu},
  journal={arXiv preprint arXiv:2502.08643},
  year={2025}
}

@article{liu2025kuda,
  title={KUDA: Keypoints to Unify Dynamics Learning and Visual Prompting for Open-Vocabulary Robotic Manipulation},
  author={Liu, Zixian and Zhang, Mingtong and Li, Yunzhu},
  journal={arXiv preprint arXiv:2503.10546},
  year={2025}
}

@inproceedings{tian2024robokeygen,
  title={Robokeygen: robot pose and joint angles estimation via diffusion-based 3D keypoint generation},
  author={Tian, Yang and Zhang, Jiyao and Huang, Guowei and Wang, Bin and Wang, Ping and Pang, Jiangmiao and Dong, Hao},
  booktitle={2024 IEEE International Conference on Robotics and Automation (ICRA)},
  pages={5375--5381},
  year={2024},
  organization={IEEE}
}

@inproceedings{simard2003best,
  title={Best practices for convolutional neural networks applied to visual document analysis.},
  author={Simard, Patrice Y and Steinkraus, David and Platt, John C and others},
  booktitle={Icdar},
  volume={3},
  number={2003},
  year={2003},
  organization={Edinburgh}
}

@article{ren2025motion,
  title={Motion tracks: A unified representation for human-robot transfer in few-shot imitation learning},
  author={Ren, Juntao and Sundaresan, Priya and Sadigh, Dorsa and Choudhury, Sanjiban and Bohg, Jeannette},
  journal={arXiv preprint arXiv:2501.06994},
  year={2025}
}

@inproceedings{li2022blip,
  title={Blip: Bootstrapping language-image pre-training for unified vision-language understanding and generation},
  author={Li, Junnan and Li, Dongxu and Xiong, Caiming and Hoi, Steven},
  booktitle={International conference on machine learning},
  pages={12888--12900},
  year={2022},
  organization={PMLR}
}

@inproceedings{xiao2024florence,
  title={Florence-2: Advancing a unified representation for a variety of vision tasks},
  author={Xiao, Bin and Wu, Haiping and Xu, Weijian and Dai, Xiyang and Hu, Houdong and Lu, Yumao and Zeng, Michael and Liu, Ce and Yuan, Lu},
  booktitle={Proceedings of the IEEE/CVF Conference on Computer Vision and Pattern Recognition},
  pages={4818--4829},
  year={2024}
}

@article{wei2022chain,
  title={Chain-of-thought prompting elicits reasoning in large language models},
  author={Wei, Jason and Wang, Xuezhi and Schuurmans, Dale and Bosma, Maarten and Xia, Fei and Chi, Ed and Le, Quoc V and Zhou, Denny and others},
  journal={Advances in neural information processing systems},
  volume={35},
  pages={24824--24837},
  year={2022}
}

@article{shorten2021text,
  title={Text data augmentation for deep learning},
  author={Shorten, Connor and Khoshgoftaar, Taghi M and Furht, Borko},
  journal={Journal of big Data},
  volume={8},
  number={1},
  pages={101},
  year={2021},
  publisher={Springer}
}

@inproceedings{kafle2017data,
  title={Data augmentation for visual question answering},
  author={Kafle, Kushal and Yousefhussien, Mohammed and Kanan, Christopher},
  booktitle={Proceedings of the 10th international conference on natural language generation},
  pages={198--202},
  year={2017}
}

@inproceedings{huang2025match,
  title={Match policy: A simple pipeline from point cloud registration to manipulation policies},
  author={Huang, Haojie and Liu, Haotian and Wang, Dian and Walters, Robin and Platt, Robert},
  booktitle={2025 IEEE International Conference on Robotics and Automation (ICRA)},
  pages={16907--16914},
  year={2025},
  organization={IEEE}
}

@article{huang2024imagination,
  title={Imagination policy: Using generative point cloud models for learning manipulation policies},
  author={Huang, Haojie and Schmeckpeper, Karl and Wang, Dian and Biza, Ondrej and Qian, Yaoyao and Liu, Haotian and Jia, Mingxi and Platt, Robert and Walters, Robin},
  journal={arXiv preprint arXiv:2406.11740},
  year={2024}
}

@article{he2021efficient,
  title={Efficient equivariant network},
  author={He, Lingshen and Chen, Yuxuan and Dong, Yiming and Wang, Yisen and Lin, Zhouchen and others},
  journal={Advances in Neural Information Processing Systems},
  volume={34},
  pages={5290--5302},
  year={2021}
}

@article{zhao2024mathrm,
  title={E(2) Equivariant Graph Planning for Navigation},
  author={Zhao, Linfeng and Li, Hongyu and Pad{\i}r, Ta{\c{s}}k{\i}n and Jiang, Huaizu and Wong, Lawson LS},
  journal={IEEE Robotics and Automation Letters},
  year={2024},
  publisher={IEEE}
}

@article{zhao2022integrating,
  title={Integrating symmetry into differentiable planning with steerable convolutions},
  author={Zhao, Linfeng and Zhu, Xupeng and Kong, Lingzhi and Walters, Robin and Wong, Lawson LS},
  journal={arXiv preprint arXiv:2206.03674},
  year={2022}
}

@inproceedings{simeonov2023se,
  title={Se (3)-equivariant relational rearrangement with neural descriptor fields},
  author={Simeonov, Anthony and Du, Yilun and Lin, Yen-Chen and Garcia, Alberto Rodriguez and Kaelbling, Leslie Pack and Lozano-P{\'e}rez, Tom{\'a}s and Agrawal, Pulkit},
  booktitle={Conference on Robot Learning},
  pages={835--846},
  year={2023},
  organization={PMLR}
}

@article{eisner2024deep,
  title={Deep SE (3)-Equivariant Geometric Reasoning for Precise Placement Tasks},
  author={Eisner, Ben and Yang, Yi and Davchev, Todor and Vecerik, Mel and Scholz, Jonathan and Held, David},
  journal={arXiv preprint arXiv:2404.13478},
  year={2024}
}

@inproceedings{radford2021learning,
  title={Learning transferable visual models from natural language supervision},
  author={Radford, Alec and Kim, Jong Wook and Hallacy, Chris and Ramesh, Aditya and Goh, Gabriel and Agarwal, Sandhini and Sastry, Girish and Askell, Amanda and Mishkin, Pamela and Clark, Jack and others},
  booktitle={International conference on machine learning},
  pages={8748--8763},
  year={2021},
  organization={PMLR}
}

@article{ze20243d,
  title={3d diffusion policy},
  author={Ze, Yanjie and Zhang, Gu and Zhang, Kangning and Hu, Chenyuan and Wang, Muhan and Xu, Huazhe},
  journal={arXiv preprint arXiv:2403.03954},
  year={2024}
}

@inproceedings{huang2023edge,
  title={Edge grasp network: A graph-based se (3)-invariant approach to grasp detection},
  author={Huang, Haojie and Wang, Dian and Zhu, Xupeng and Walters, Robin and Platt, Robert},
  booktitle={2023 IEEE International Conference on Robotics and Automation (ICRA)},
  pages={3882--3888},
  year={2023},
  organization={IEEE}
}

@inproceedings{simeonov2022neural,
  title={Neural descriptor fields: Se (3)-equivariant object representations for manipulation},
  author={Simeonov, Anthony and Du, Yilun and Tagliasacchi, Andrea and Tenenbaum, Joshua B and Rodriguez, Alberto and Agrawal, Pulkit and Sitzmann, Vincent},
  booktitle={2022 International Conference on Robotics and Automation (ICRA)},
  pages={6394--6400},
  year={2022},
  organization={IEEE}
}

@inproceedings{pan2023tax,
  title={Tax-pose: Task-specific cross-pose estimation for robot manipulation},
  author={Pan, Chuer and Okorn, Brian and Zhang, Harry and Eisner, Ben and Held, David},
  booktitle={Conference on Robot Learning},
  pages={1783--1792},
  year={2023},
  organization={PMLR}
}

@article{chi2023diffusion,
  title={Diffusion policy: Visuomotor policy learning via action diffusion},
  author={Chi, Cheng and Feng, Siyuan and Du, Yilun and Xu, Zhenjia and Cousineau, Eric and Burchfiel, Benjamin and Song, Shuran},
  journal={arXiv preprint arXiv:2303.04137},
  year={2023}
}

@article{song2020denoising,
  title={Denoising diffusion implicit models},
  author={Song, Jiaming and Meng, Chenlin and Ermon, Stefano},
  journal={arXiv preprint arXiv:2010.02502},
  year={2020}
}

@inproceedings{
        huang2024fourier,
        title={Fourier Transporter: Bi-Equivariant Robotic Manipulation in 3D},
        author={Haojie Huang and Owen Lewis Howell and Dian Wang and Xupeng Zhu and Robert Platt and Robin Walters},
        booktitle={The Twelfth International Conference on Learning Representations},
        year={2024},
        url={https://openreview.net/forum?id=UulwvAU1W0}}

@article{huang2024leveraging,
  title={Leveraging symmetries in pick and place},
  author={Huang, Haojie and Wang, Dian and Tangri, Arsh and Walters, Robin and Platt, Robert},
  journal={The International Journal of Robotics Research},
  pages={02783649231225775},
  year={2024},
  publisher={SAGE Publications Sage UK: London, England}
}

@INPROCEEDINGS{Huang-RSS-22, 
    AUTHOR    = {Haojie Huang AND Dian Wang AND Robin Walters AND Robert Platt}, 
    TITLE     = {{Equivariant Transporter Network}}, 
    BOOKTITLE = {Proceedings of Robotics: Science and Systems}, 
    YEAR      = {2022}, 
    ADDRESS   = {New York City, NY, USA}, 
    MONTH     = {June}, 
    DOI       = {10.15607/RSS.2022.XVIII.007} 
}

@article{zhu2022grasp,
  title={Sample Efficient Grasp Learning Using Equivariant Models},
  author={Zhu, Xupeng and Wang, Dian and Biza, Ondrej and Su, Guanang and Walters, Robin and Platt, Robert},
  journal={Proceedings of Robotics: Science and Systems (RSS)},
  year={2022} }

@inproceedings{jia2023seil,
  title={SEIL: Simulation-augmented Equivariant Imitation Learning},
  author={Jia, Mingxi and Wang, Dian and Su, Guanang and Klee, David and Zhu, Xupeng and Walters, Robin and Platt, Robert},
  booktitle={2023 IEEE International Conference on Robotics and Automation (ICRA)},
  pages={1845--1851},
  year={2023},
  organization={IEEE}
}

@inproceedings{
wang2022so2equivariant,
title={{$\mathrm{SO}(2)$}-Equivariant Reinforcement Learning},
author={Dian Wang and Robin Walters and Robert Platt},
booktitle={International Conference on Learning Representations},
year={2022},
url={https://openreview.net/forum?id=7F9cOhdvfk_}
}

@inproceedings{
wang2022onrobot,
title={On-Robot Learning With Equivariant Models},
author={Dian Wang and Mingxi Jia and Xupeng Zhu and Robin Walters and Robert Platt},
booktitle={6th Annual Conference on Robot Learning},
year={2022},
url={https://openreview.net/forum?id=K8W6ObPZQyh}
}

@article{ryu2022equivariant,
  title={Equivariant descriptor fields: Se (3)-equivariant energy-based models for end-to-end visual robotic manipulation learning},
  author={Ryu, Hyunwoo and Lee, Hong-in and Lee, Jeong-Hoon and Choi, Jongeun},
  journal={arXiv preprint arXiv:2206.08321},
  year={2022}
}

@article{ryu2023diffusion,
  title={Diffusion-edfs: Bi-equivariant denoising generative modeling on se (3) for visual robotic manipulation},
  author={Ryu, Hyunwoo and Kim, Jiwoo and Chang, Junwoo and Ahn, Hyun Seok and Seo, Joohwan and Kim, Taehan and Choi, Jongeun and Horowitz, Roberto},
  journal={arXiv preprint arXiv:2309.02685},
  year={2023}
}

@inproceedings{e2cnn,
    title={{General E(2)-Equivariant Steerable CNNs}},
    author={Weiler, Maurice and Cesa, Gabriele},
    booktitle={Conference on Neural Information Processing Systems (NeurIPS)},
    year={2019},
}

@inproceedings{cesa2022a,
        title={A Program to Build {E(N)}-Equivariant Steerable {CNN}s },
        author={Gabriele Cesa and Leon Lang and Maurice Weiler},
        booktitle={International Conference on Learning Representations},
        year={2022},
        url={https://openreview.net/forum?id=WE4qe9xlnQw}
    }

@inproceedings{deng2021vector,
  title={Vector neurons: A general framework for so (3)-equivariant networks},
  author={Deng, Congyue and Litany, Or and Duan, Yueqi and Poulenard, Adrien and Tagliasacchi, Andrea and Guibas, Leonidas J},
  booktitle={Proceedings of the IEEE/CVF International Conference on Computer Vision},
  pages={12200--12209},
  year={2021}
}

@article{liao2022equiformer,
  title={Equiformer: Equivariant graph attention transformer for 3d atomistic graphs},
  author={Liao, Yi-Lun and Smidt, Tess},
  journal={arXiv preprint arXiv:2206.11990},
  year={2022}
}

@inproceedings{huorbitgrasp,
  title={OrbitGrasp: {SE} (3)-Equivariant Grasp Learning},
  author={Hu, Boce and Zhu, Xupeng and Wang, Dian and Dong, Zihao and Huang, Haojie and Wang, Chenghao and Walters, Robin and Platt, Robert},
    year={2024},
  booktitle={8th Annual Conference on Robot Learning}
}

@article{wang2024equivariant,
  title={Equivariant Diffusion Policy},
  author={Wang, Dian and Hart, Stephen and Surovik, David and Kelestemur, Tarik and Huang, Haojie and Zhao, Haibo and Yeatman, Mark and Wang, Jiuguang and Walters, Robin and Platt, Robert},
  journal={arXiv preprint arXiv:2407.01812},
  year={2024}
}

@inproceedings{yang2024equivact,
  title={Equivact: Sim (3)-equivariant visuomotor policies beyond rigid object manipulation},
  author={Yang, Jingyun and Deng, Congyue and Wu, Jimmy and Antonova, Rika and Guibas, Leonidas and Bohg, Jeannette},
  booktitle={2024 IEEE International Conference on Robotics and Automation (ICRA)},
  pages={9249--9255},
  year={2024},
  organization={IEEE}
}

@inproceedings{wang2021q,
	author = {Dian Wang and Robin Walters and Xupeng Zhu and Robert Platt},
	booktitle = {5th Annual Conference on Robot Learning},
	title = {{Equivariant {$Q$} Learning in Spatial Action Spaces}},
	year = {2021}}

@inproceedings{liu2023continual,
	author = {Liu, Shiqi and Xu, Mengdi and Huang, Peide and Zhang, Xilun and Liu, Yongkang and Oguchi, Kentaro and Zhao, Ding},
	booktitle = {Conference on Robot Learning},
	organization = {PMLR},
	pages = {222--240},
	title = {{Continual Vision-based Reinforcement Learning with Group Symmetries}},
	year = {2023}}

@article{kohler2023symmetric,
  title={Symmetric models for visual force policy learning},
  author={Kohler, Colin and Srikanth, Anuj Shrivatsav and Arora, Eshan and Platt, Robert},
  journal={arXiv preprint arXiv:2308.14670},
  year={2023}
}

@inproceedings{nguyen2023equivariant,
  title={Equivariant reinforcement learning under partial observability},
  author={Nguyen, Hai Huu and Baisero, Andrea and Klee, David and Wang, Dian and Platt, Robert and Amato, Christopher},
  booktitle={Conference on Robot Learning},
  pages={3309--3320},
  year={2023},
  organization={PMLR}
}

@article{nguyen2024symmetry,
  title={Symmetry-aware Reinforcement Learning for Robotic Assembly under Partial Observability with a Soft Wrist},
  author={Nguyen, Hai and Kozuno, Tadashi and Beltran-Hernandez, Cristian C and Hamaya, Masashi},
  journal={arXiv preprint arXiv:2402.18002},
  year={2024}
}

@article{yang2024equibot,
  title={EquiBot: SIM (3)-Equivariant Diffusion Policy for Generalizable and Data Efficient Learning},
  author={Yang, Jingyun and Cao, Zi-ang and Deng, Congyue and Antonova, Rika and Song, Shuran and Bohg, Jeannette},
  journal={arXiv preprint arXiv:2407.01479},
  year={2024}
}

@article{zhu2023robot,
  title={On robot grasp learning using equivariant models},
  author={Zhu, Xupeng and Wang, Dian and Su, Guanang and Biza, Ondrej and Walters, Robin and Platt, Robert},
  journal={Autonomous Robots},
  volume={47},
  number={8},
  pages={1175--1193},
  year={2023},
  publisher={Springer}
}

@article{haldar2026point,
  title={Point Bridge: 3D Representations for Cross Domain Policy Learning},
  author={Haldar, Siddhant and Johannsmeier, Lars and Pinto, Lerrel and Gupta, Abhishek and Fox, Dieter and Narang, Yashraj and Mandlekar, Ajay},
  journal={arXiv preprint arXiv:2601.16212},
  year={2026}
}

@inproceedings{hamdi2021mvtn,
  title={Mvtn: Multi-view transformation network for 3d shape recognition},
  author={Hamdi, Abdullah and Giancola, Silvio and Ghanem, Bernard},
  booktitle={Proceedings of the IEEE/CVF international conference on computer vision},
  pages={1--11},
  year={2021}
}

\clearpage
\section{Appendix}


\subsection{Multi-task Extension and Comparison}
\label{multi-task-extension}

We also consider a multi-task evaluation setting to assess the capacity of our method when a single policy is trained jointly across all tasks (note that multi-task learning is orthogonal to our main contribution). Task instructions are embedded into fixed-length token sequences via a CLIP encoder and fed into the cross-attention layers of the multi-view Transformer.

We compare against two baselines: (1) \textbf{LBM}~\cite{barreiros2025careful}, which extends Diffusion Policy to the multi-task setting using pretrained CLIP~\cite{radford2021learning} visual and language encoders. As it is not open-sourced, we reimplement the architecture following the paper. (2) \textbf{ManiFlow}~\cite{yan2025maniflow}, which uses a diffusion Transformer based on consistency flow matching~\cite{lipman2022flow}.

\begin{table*}[ht]
  \setlength\tabcolsep{1.5pt}
  \begin{center}
  \scriptsize
  \begin{tabular}{@{}lcccccccccccc@{}}
  \toprule
  & stack-three-d1 & hammer-cleanup-d1 & mug-cleanup-d1 & coffee-d2 & square-d2 & threading-d2 & average score \\
  \cmidrule(lr){2-2} \cmidrule(lr){3-3} \cmidrule(lr){4-4} \cmidrule(lr){5-5} \cmidrule(lr){6-6} \cmidrule(lr){7-7} \cmidrule(lr){8-8} 
  Method & 100 Demo & 100 Demo & 100 Demo & 100 Demo & 100 Demo & 100 Demo & 600 Demo (total)\\
  \midrule
  LBM~\cite{barreiros2025careful} & 86 & 84 & 66 & 66 & 62 & 38 & 67.0 \\
  Maniflow~\cite{yan2025maniflow} & 30 & 70 & 38 & 44 & 8 & 12 & 33.7   \\
  $\ours$ - multiple task & 94 & 88 & 72 & 60 & 58 & 50 & 70.3\\
  $\ours$ - single task & 94 & 81 & 79 & 62 & 50 & 52 & 69.7 \\
  \bottomrule
  \end{tabular}
  \end{center}
  \vspace{-0.5em}
  \caption{\scriptsize\textbf{Performance Comparisons on Mimicgen Benchmark - Multi-task Setting.} Success rate (mean\%) on 50 unseen tests with 600 demonstration episodes used for training.}
  \label{table:multi-task}
\end{table*}

\begin{figure*}[ht]
  \centering
  \includegraphics[width=0.7\textwidth]{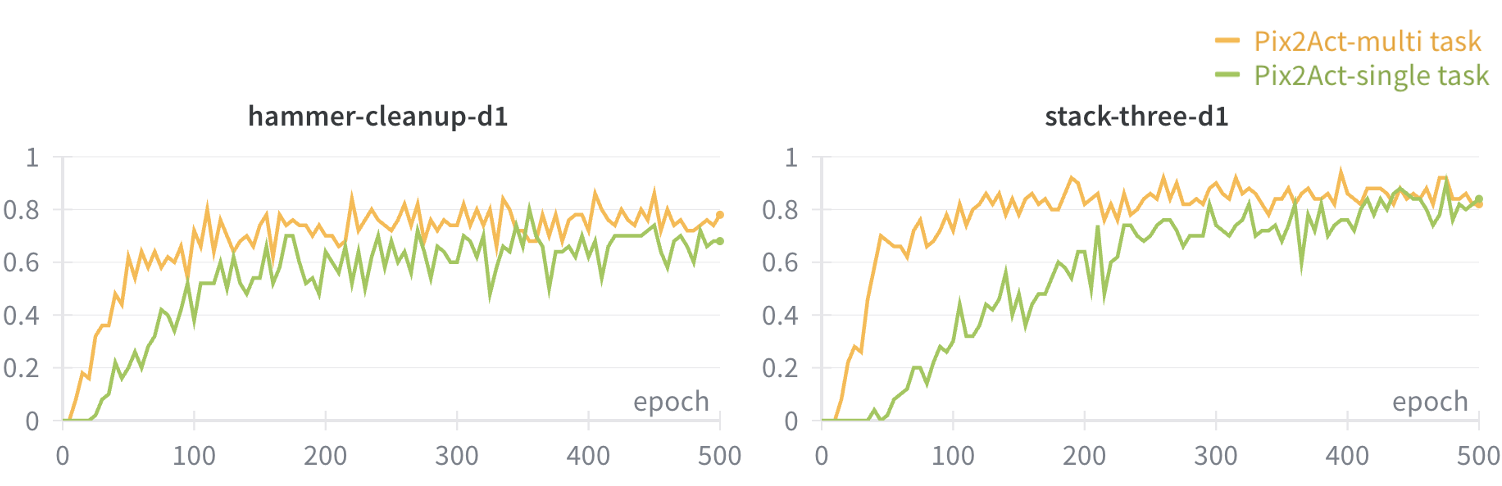}
  \caption{\textbf{Multi-task Training Demonstrates Faster Convergence across Epochs.} Test Success Rate as a function of Training Epoch.}
  \label{fig:table2-figure}
\end{figure*}
As shown in Table~\ref{table:multi-task}, $\ours$ outperforms two strong multi-task baselines. LBM achieves slightly lower performance than $\ours$ despite using a large pretrained encoder with roughly five times more parameters, suggesting that the inductive biases of our image-based action representation are more effective than scale alone for this setting. While multi-task training does not yield significant performance gains over our single-task results, it produces noticeably faster convergence across epochs (Figure~\ref{fig:table2-figure}). Combining $\ours$ with a stronger pretrained vision encoder is a natural direction for future work.

\subsection{Illustration of $n$-Equivariance}
\label{appendix-equivariance}

The two propositions together define the \textit{$n$-equivariance} symmetry of the policy:
\[
\pi(G \cdot O) = G \cdot \pi(O), \qquad G \in \SE(2)^n \rtimes S_n,
\]
where $G$ combines $n$ independent planar rigid motions with permutations of the $n$ cameras. Concretely, $G$ corresponds to rotating and translating each camera around its optical axis, or swapping cameras whose roles are interchangeable (e.g., the two in-hand views of a parallel-jaw gripper). Since these camera-side transformations are canceled out during triangulation, \textbf{the triangulated 3D policy is invariant to them}: the recovered action does not change when individual cameras are perturbed or reindexed.

This symmetry has three useful consequences.

\textbf{(1) Robustness to novel views.} Because the policy learns equivariant keypoint trajectories rather than view-specific ones, it generalizes to camera configurations not seen during training. In our real-world experiments, randomly shaking the agent-view camera at inference time leaves the policy's performance unaffected (Section~\ref{appendix-realworld}).

\textbf{(2) Symmetry under 180° gripper rotations.} For parallel-jaw grippers, rotating the gripper 180\textdegree{} around its approach axis simply swaps the two in-hand cameras, so the in-hand image pair $(o_1, o_2)$ and $(o_2, o_1)$ should produce the same 3D action. This symmetry is automatically captured by the permutation component $S_n$ of our group, factoring out a redundancy in the observation space that the policy would otherwise need to learn from data.

\textbf{(3) Combinatorial generalization across views.} Suppose the training set contains two trajectories with in-hand image pairs $(o_1^i, o_2^i)$ and $(o_1^j, o_2^j)$, producing actions $(a_1^i, a_2^i)$ and $(a_1^j, a_2^j)$ respectively. At inference, the policy may encounter a novel pair
\[
o_1^k = g_1^k \cdot o_1^i, \qquad o_2^k = g_2^k \cdot o_2^j,
\]
where $g_1^k$ and $g_2^k$ are independent per-camera transformations and the pair itself was never observed during training. By $n$-equivariance,
\[
\pi(o_1^k, o_2^k) = (g_1^k \cdot a_1^i,\; g_2^k \cdot a_2^j),
\]
so the policy can transfer manipulation knowledge by recombining transformed components from different training samples. This effectively enlarges the support of the training distribution.

\subsection{Real-World Experiment Details}
\label{appendix-realworld}

\begin{figure}[ht]
  \centering
  \begin{subfigure}[t]{0.235\textwidth}
    \includegraphics[width=\linewidth]{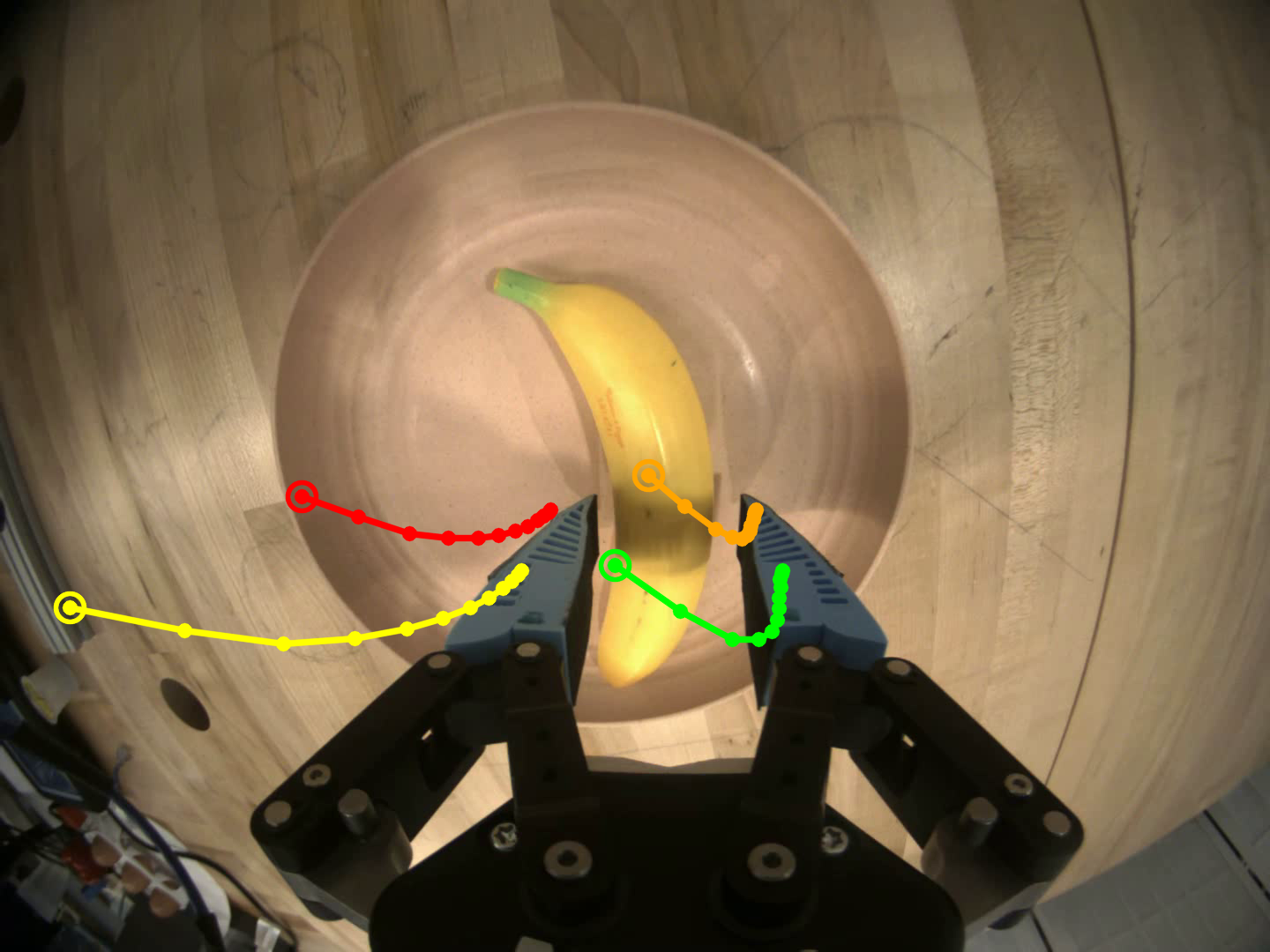}
    \caption{Prepare Fruit (top)}
  \end{subfigure}\hspace{2em}
  \begin{subfigure}[t]{0.235\textwidth}
    \includegraphics[width=\linewidth]{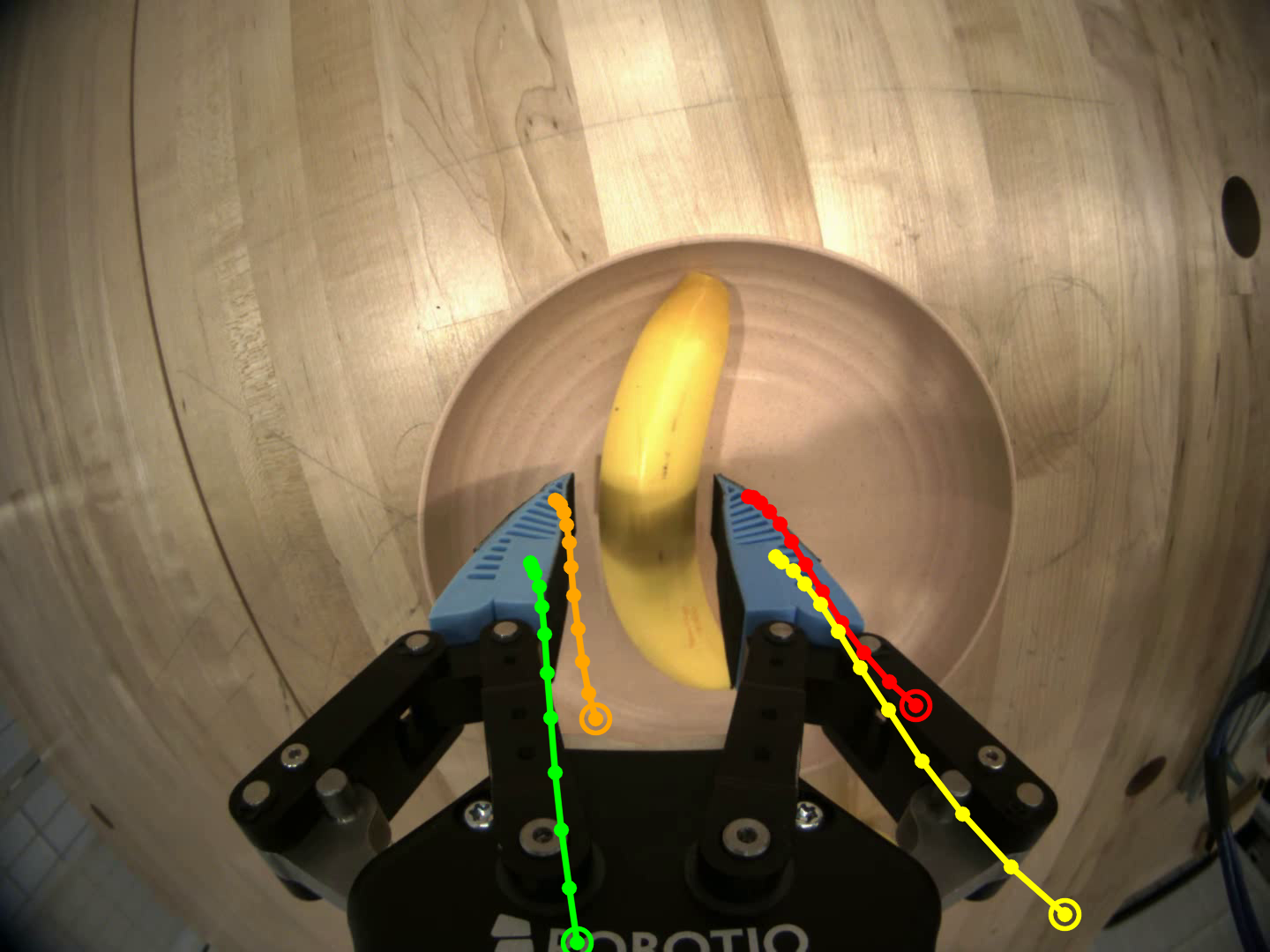}
    \caption{Prepare Fruit (bottom)}
  \end{subfigure}

  \vspace{0.4em}

  \begin{subfigure}[t]{0.235\textwidth}
    \includegraphics[width=\linewidth]{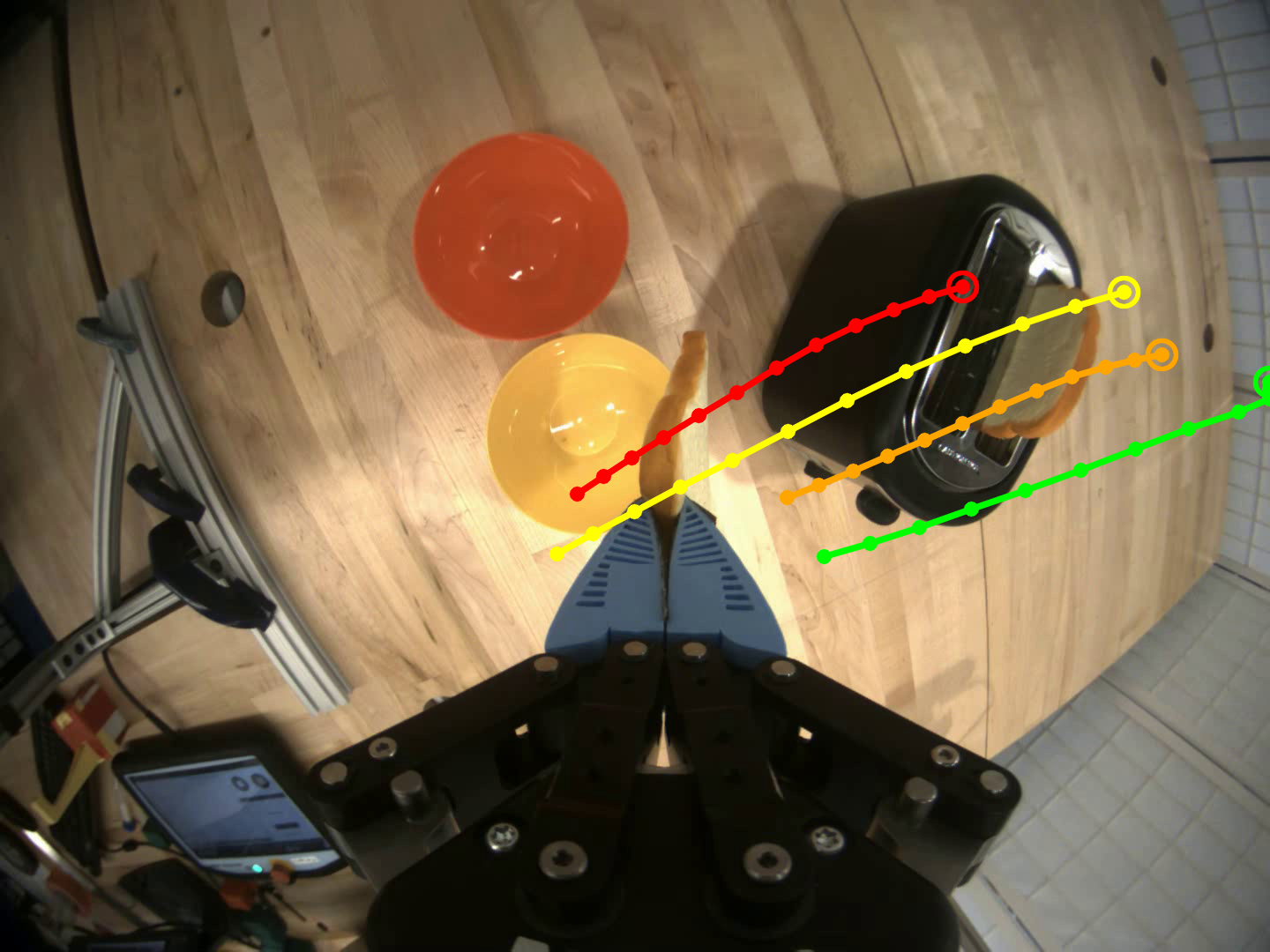}
    \caption{Make Toast (top)}
  \end{subfigure}\hspace{2em}
  \begin{subfigure}[t]{0.235\textwidth}
    \includegraphics[width=\linewidth]{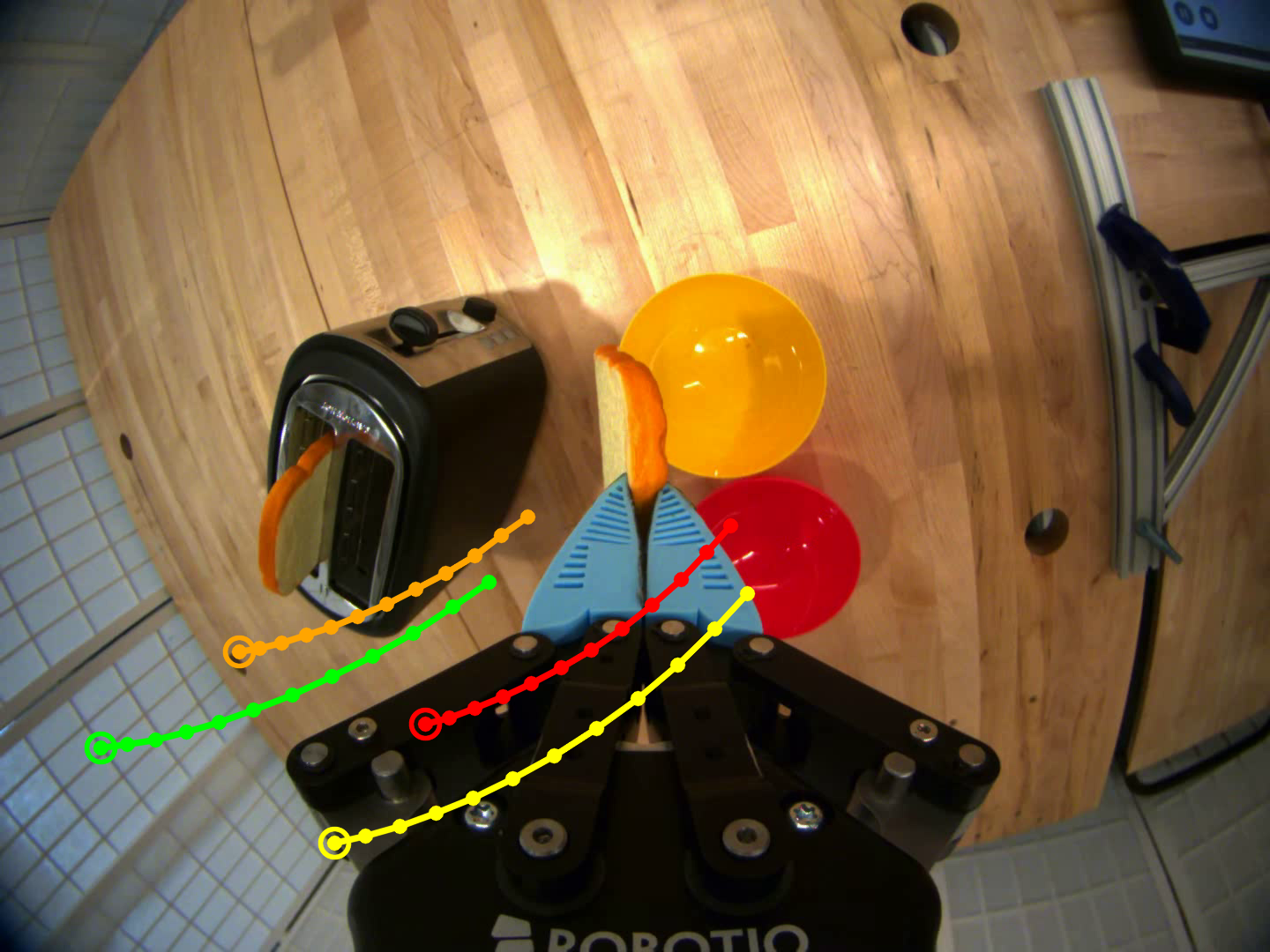}
    \caption{Make Toast (bottom)}
  \end{subfigure}
  \caption{\textbf{Predicted trajectory visualization.} Predicted image-space keypoint trajectories overlaid on the top and bottom in-hand camera views for two real-world tasks. Trajectories remain consistent across the two views.}
  \label{fig:real-traj-vis}
\end{figure}
\textbf{Visualization of predicted trajectories.}
Figure~\ref{fig:real-traj-vis} shows the predicted keypoint trajectories on the two in-hand cameras for the \textbf{Prepare Fruit Plate} and \textbf{Make Toast} tasks. The predicted trajectories remain consistent and geometrically aligned across views, confirming that the cross-view design produces coherent multi-view predictions.

\begin{figure}[h]
  \centering
  \begin{subfigure}[t]{0.31\textwidth}
    \includegraphics[width=\linewidth]{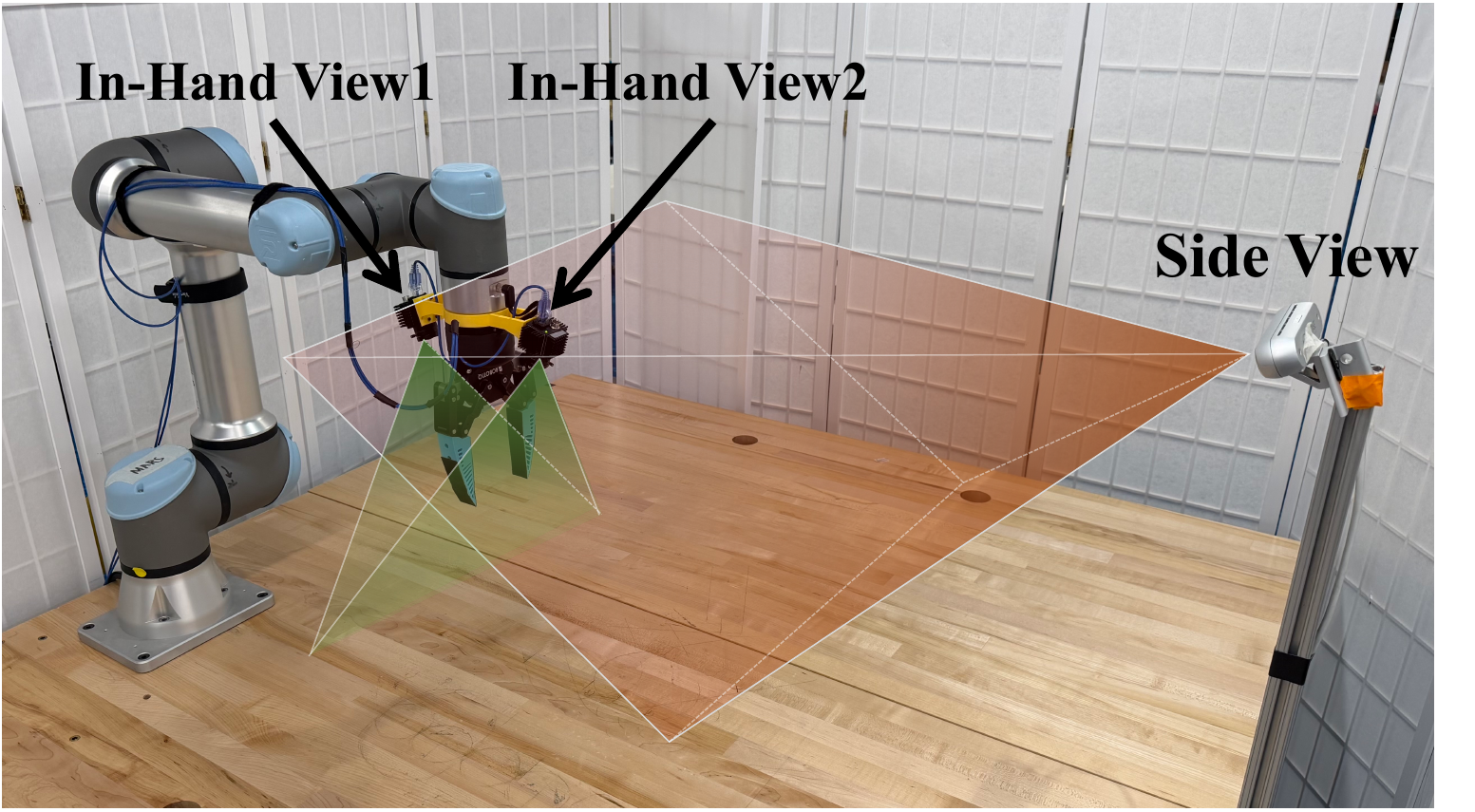}
    \caption{Robot setup.}
    \label{fig:setup-a}
  \end{subfigure}\hfill
  \begin{subfigure}[t]{0.17\textwidth}
    \includegraphics[width=\linewidth]{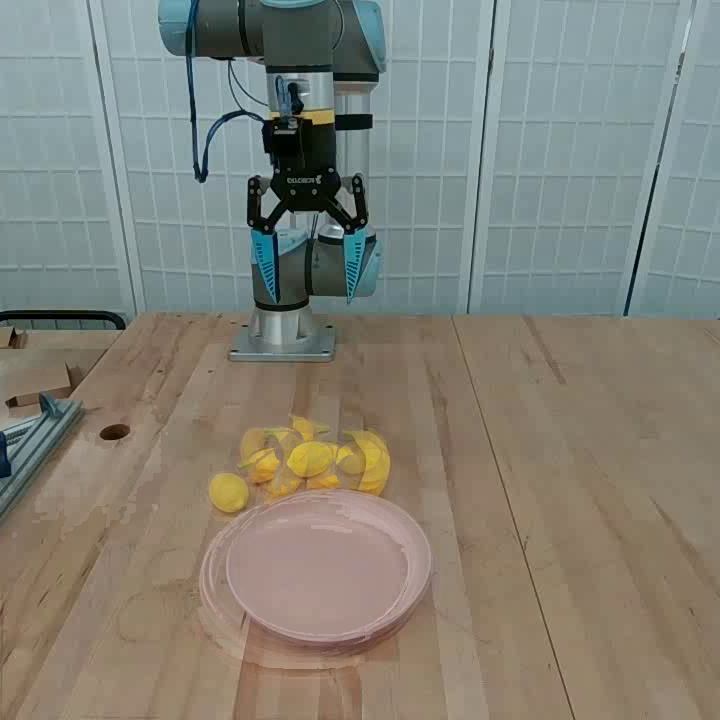}
    \caption{Initial variations for Prepare Fruit.}
    \label{fig:setup-b}
  \end{subfigure}\hfill
  \begin{subfigure}[t]{0.17\textwidth}
    \includegraphics[width=\linewidth]{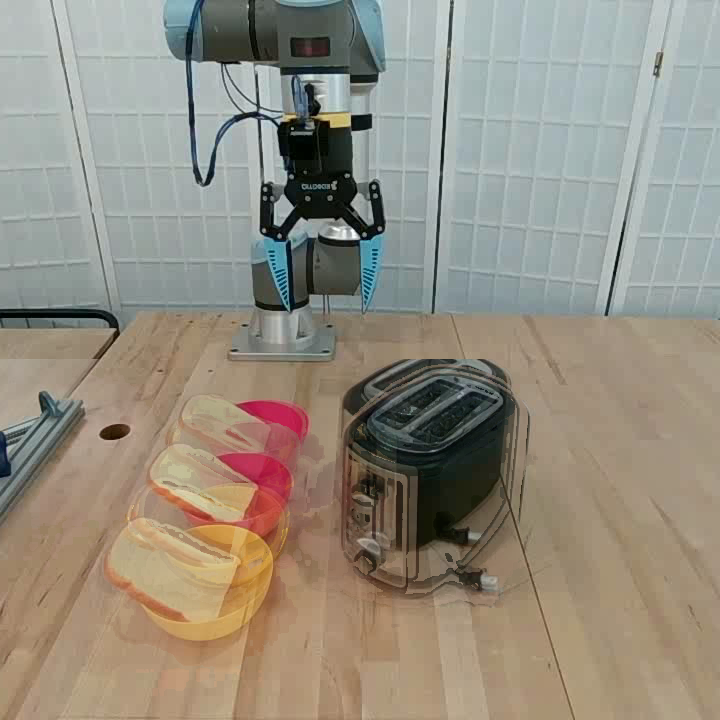}
    \caption{Initial variations for Make Toast.}
    \label{fig:setup-c}
  \end{subfigure}\hfill
  \begin{subfigure}[t]{0.17\textwidth}
    \includegraphics[width=\linewidth]{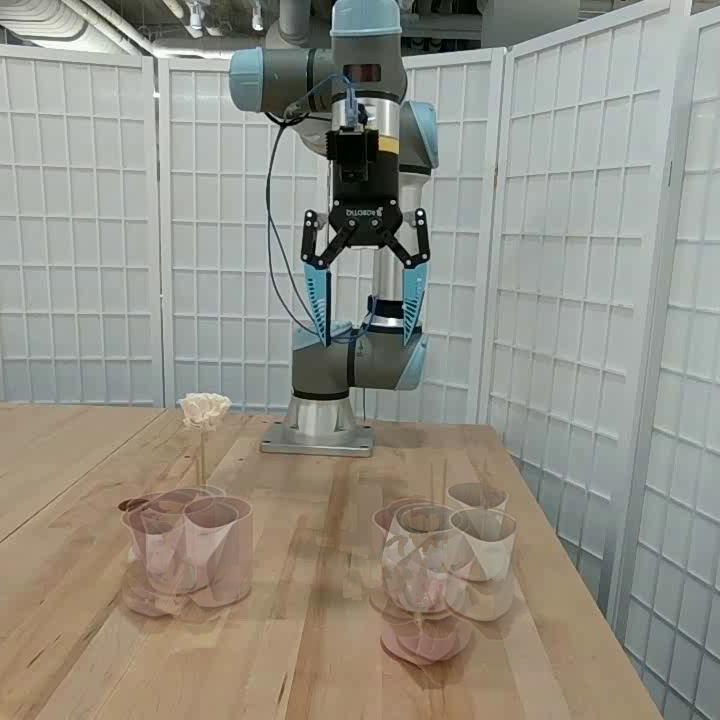}
    \caption{Initial variations for Plug Flower.}
    \label{fig:setup-d}
  \end{subfigure}\hfill
  \begin{subfigure}[t]{0.17\textwidth}
    \includegraphics[width=\linewidth]{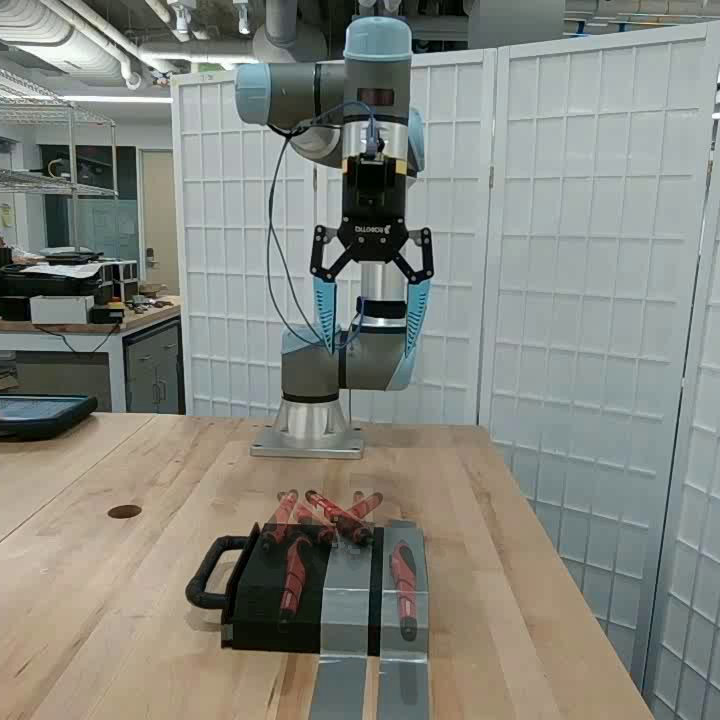}
    \caption{Initial variations for Put in Drawer.}
    \label{fig:setup-e}
  \end{subfigure}\
  \caption{\textbf{Real-world robot setup and initial task variations.} (a) The UR5 platform with two FLIR Blackfly in-hand cameras and one Intel RealSense D455 agent-view camera. (b--e) Representative initial object configurations for two of the four tasks; configurations are randomized within the workspace at the start of each trial.}
  \label{fig:real-setup}
\end{figure}

\textbf{Robot setup and task initial variations.}
We use a UR5 robotic arm equipped with two FLIR Blackfly in-hand cameras and one Intel RealSense D455 agent-view camera (Figure~\ref{fig:real-setup}, left). For each task, the initial configurations of the manipulated objects vary across trials within a fixed workspace region; representative initial configurations are shown for two tasks (Figure~\ref{fig:real-setup}, middle and right).

\begin{figure}[h]
  \centering
  \captionsetup[subfigure]{justification=centering, labelformat=empty}
  \begin{subfigure}[t]{0.192\textwidth}
    \includegraphics[width=\linewidth]{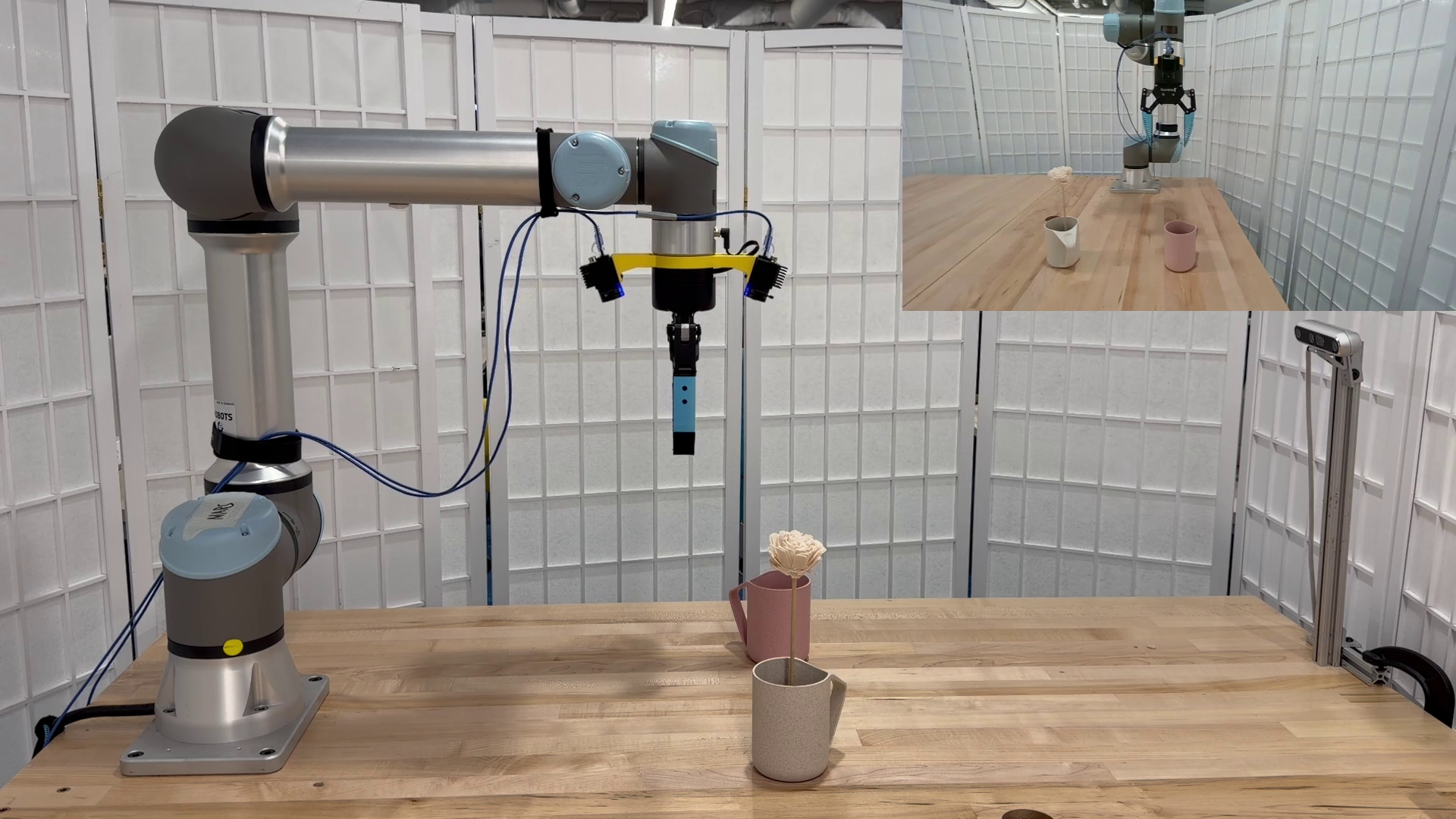}
    \caption{Plug Flower}
  \end{subfigure}\hspace{1pt}%
  \begin{subfigure}[t]{0.195\textwidth}
    \includegraphics[width=\linewidth]{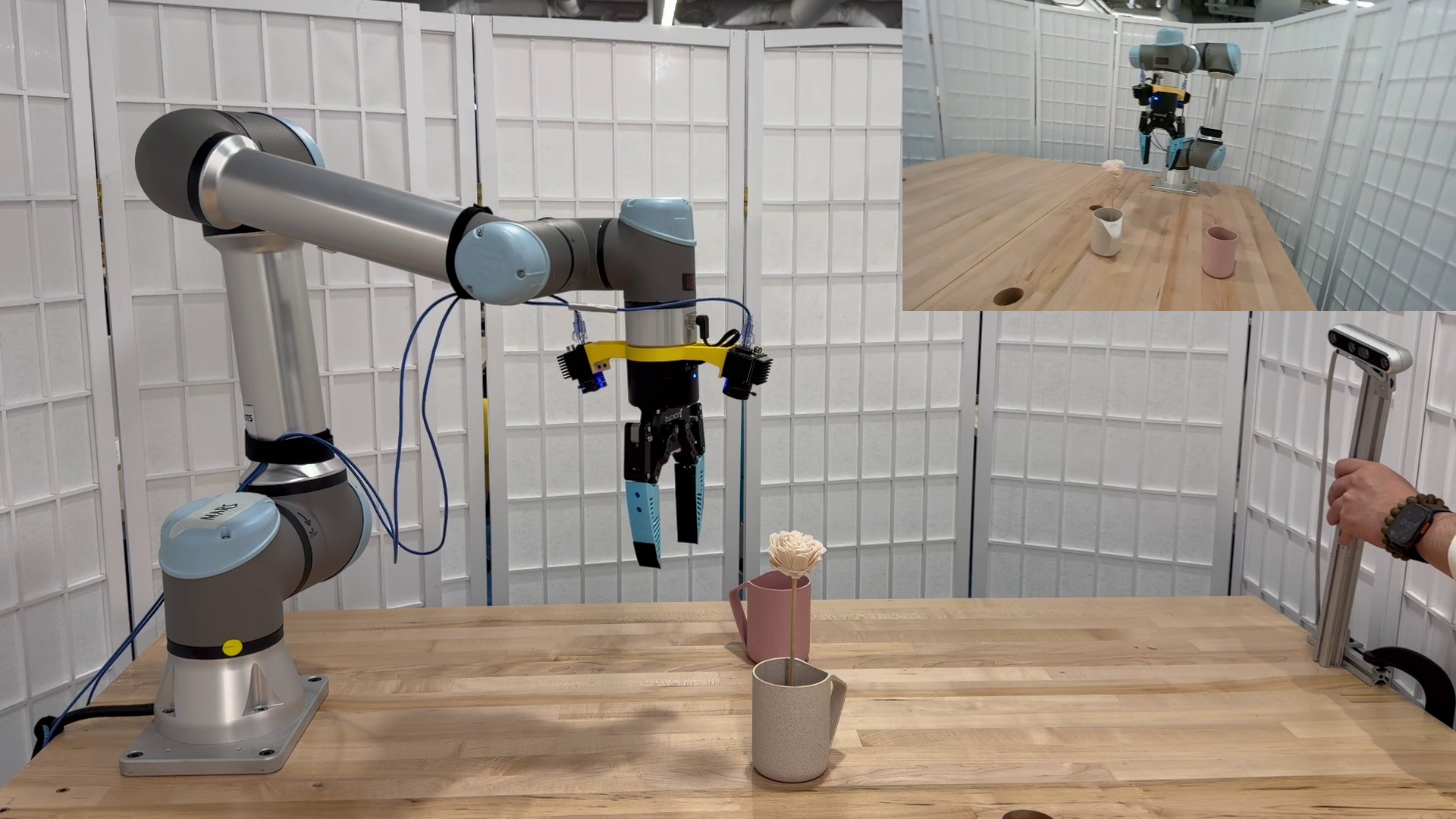}
  \end{subfigure}\hspace{1pt}%
  \begin{subfigure}[t]{0.195\textwidth}
    \includegraphics[width=\linewidth]{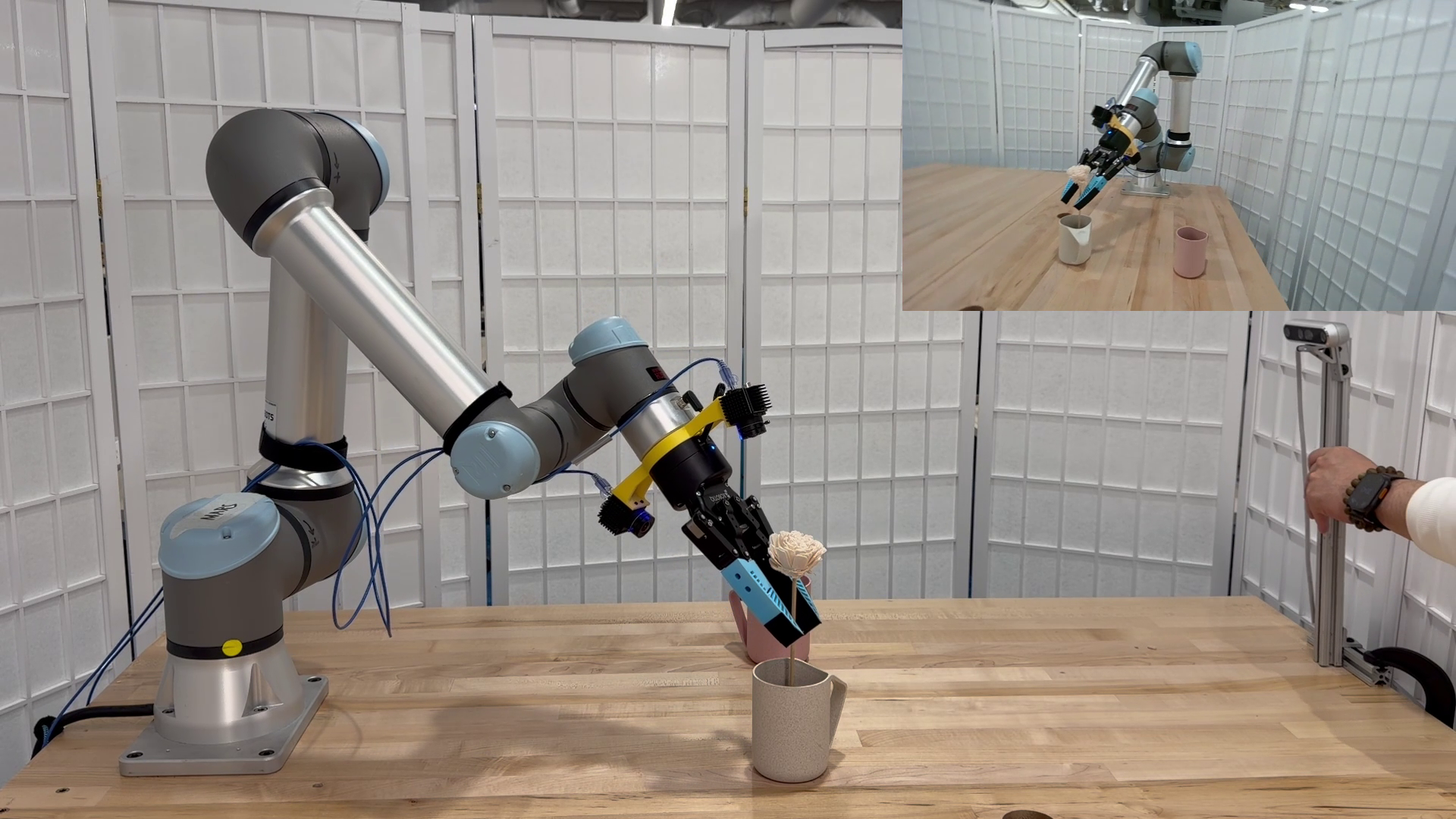}
  \end{subfigure}\hspace{1pt}%
  \begin{subfigure}[t]{0.195\textwidth}
    \includegraphics[width=\linewidth]{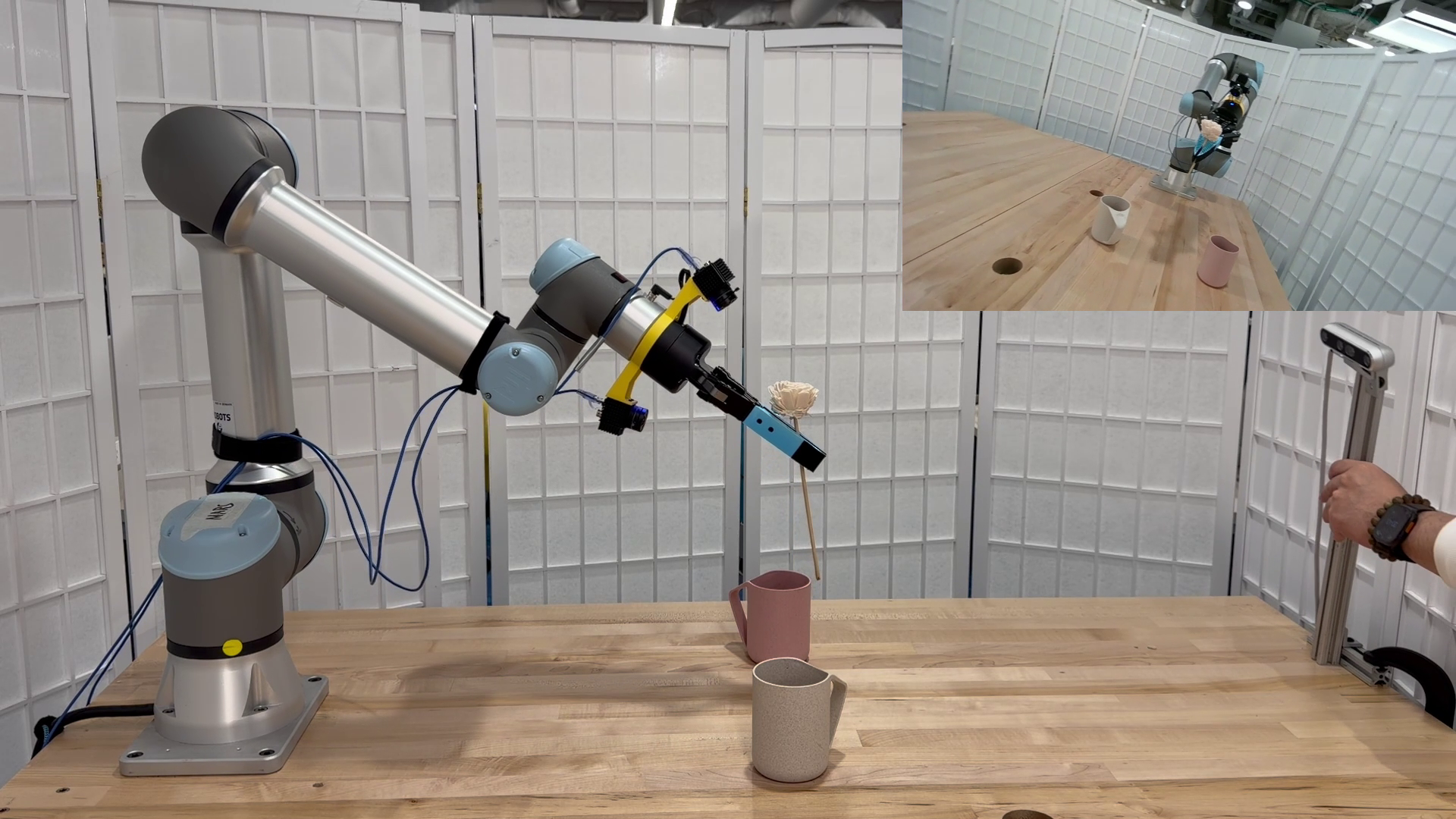}
  \end{subfigure}\hspace{1pt}%
  \begin{subfigure}[t]{0.195\textwidth}
    \includegraphics[width=\linewidth]{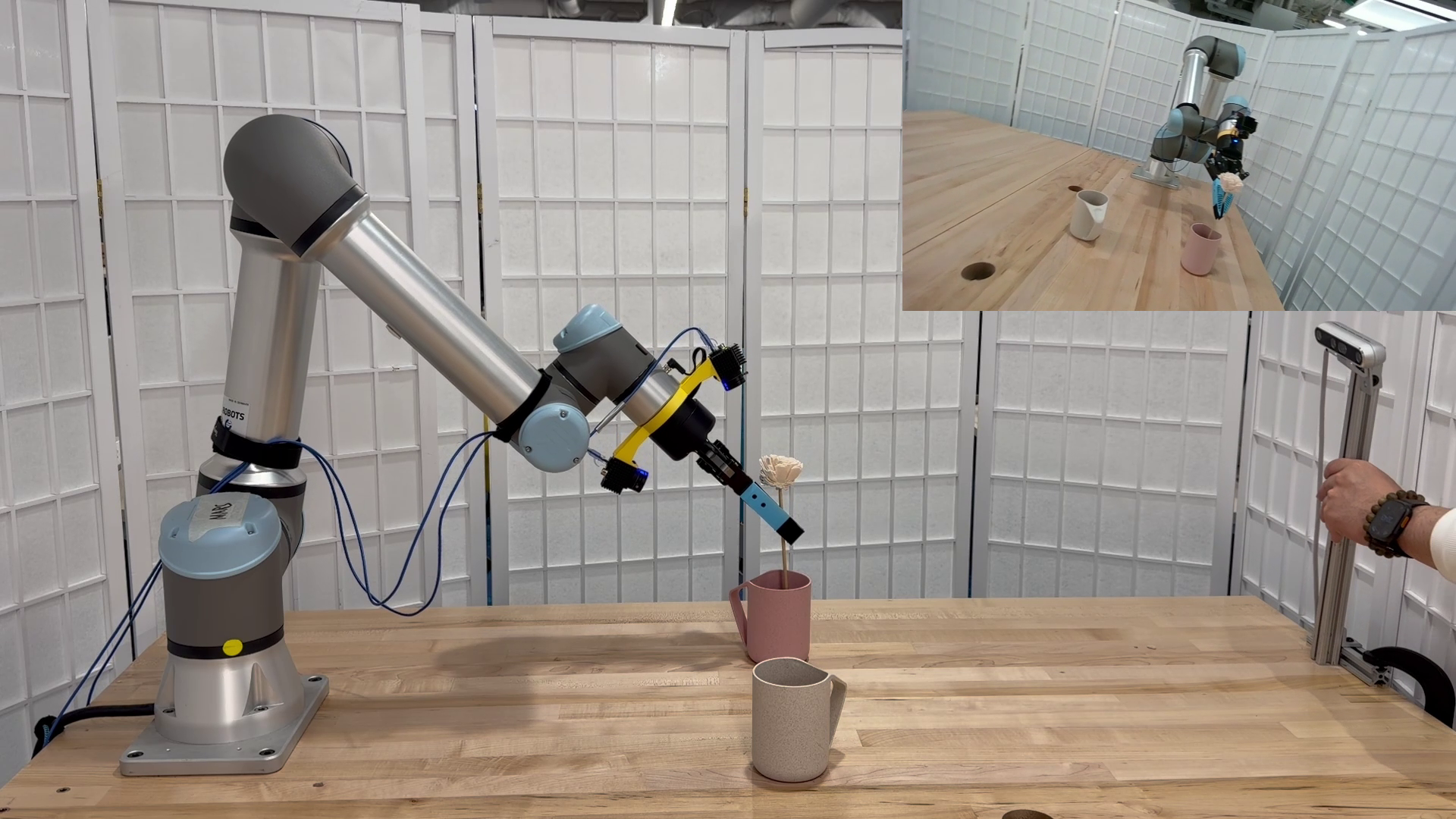}
  \end{subfigure}

  \vspace{0.4em}

  \begin{subfigure}[t]{0.195\textwidth}
    \includegraphics[width=\linewidth]{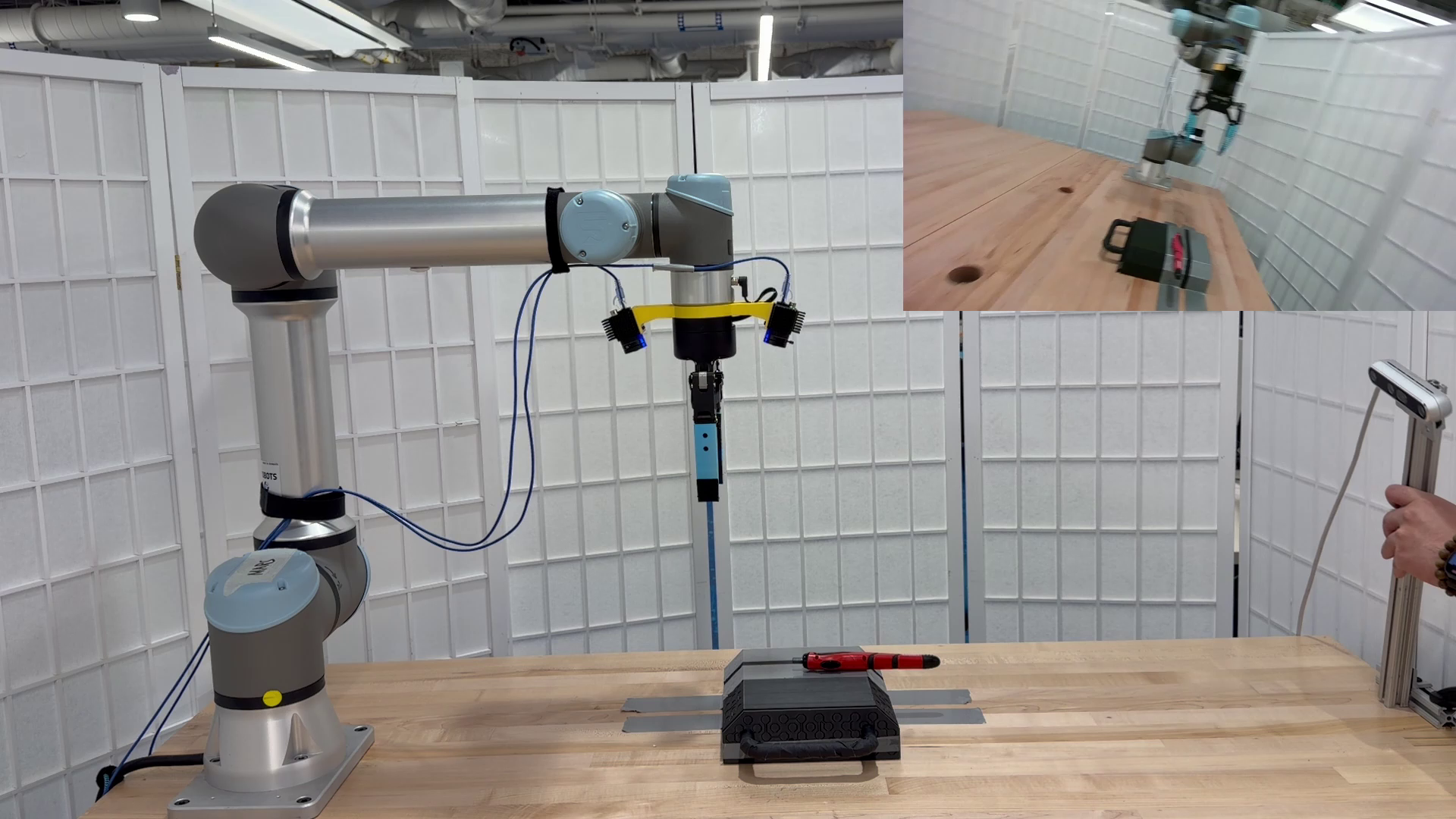}
    \caption{Put in Drawer}
  \end{subfigure}\hspace{1pt}%
  \begin{subfigure}[t]{0.195\textwidth}
    \includegraphics[width=\linewidth]{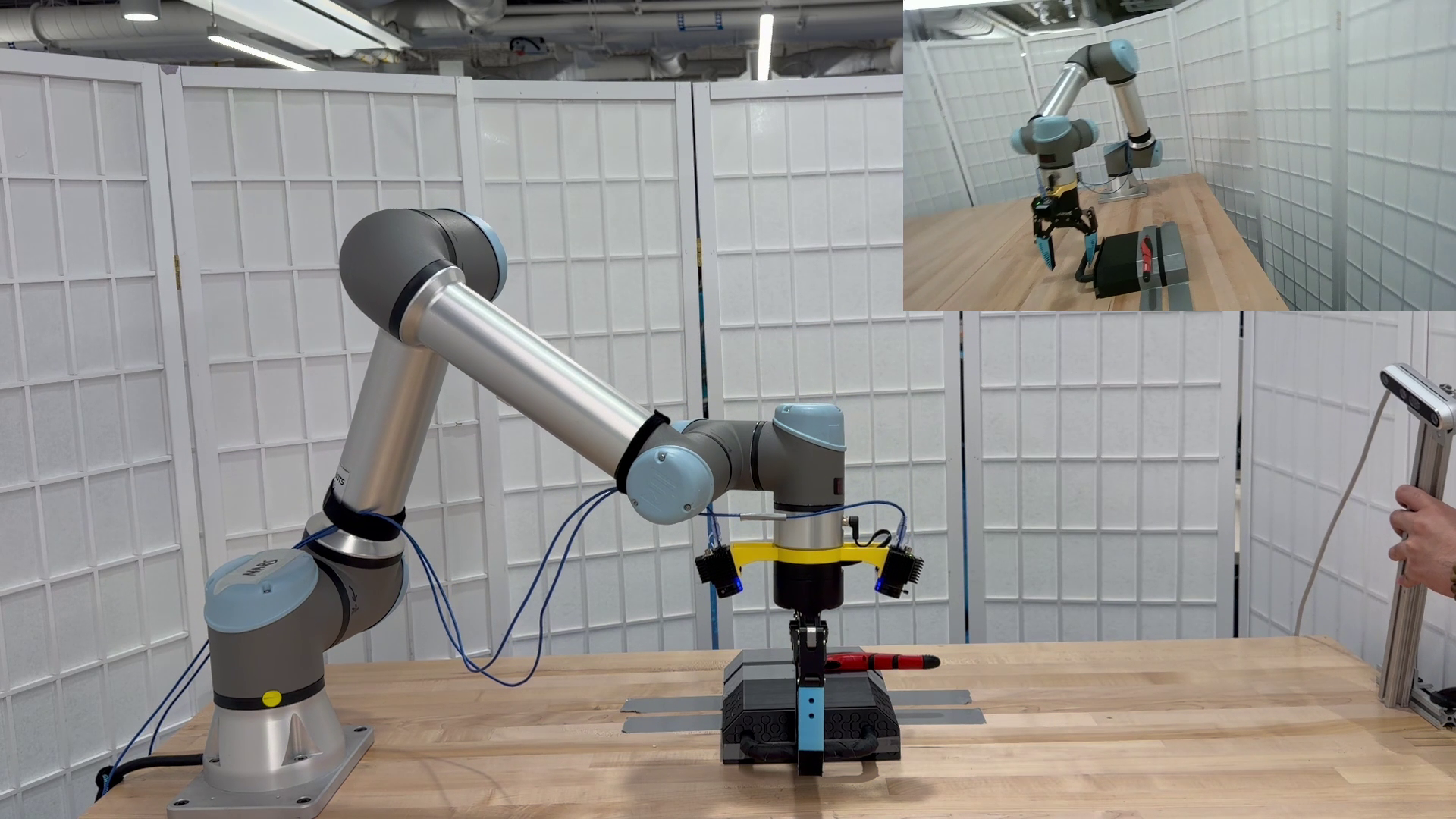}
  \end{subfigure}\hspace{1pt}%
  \begin{subfigure}[t]{0.195\textwidth}
    \includegraphics[width=\linewidth]{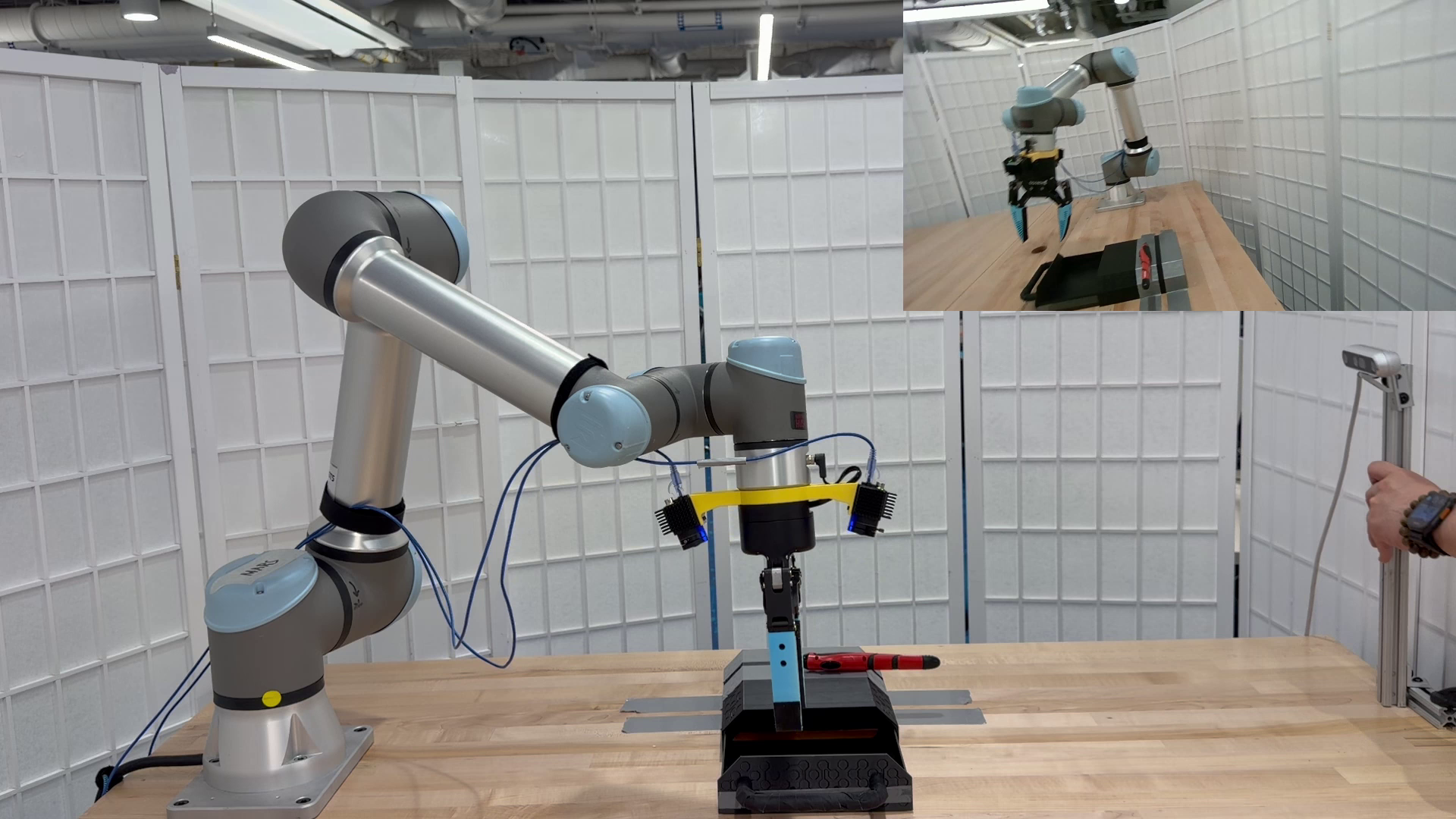}
  \end{subfigure}\hspace{1pt}%
  \begin{subfigure}[t]{0.195\textwidth}
    \includegraphics[width=\linewidth]{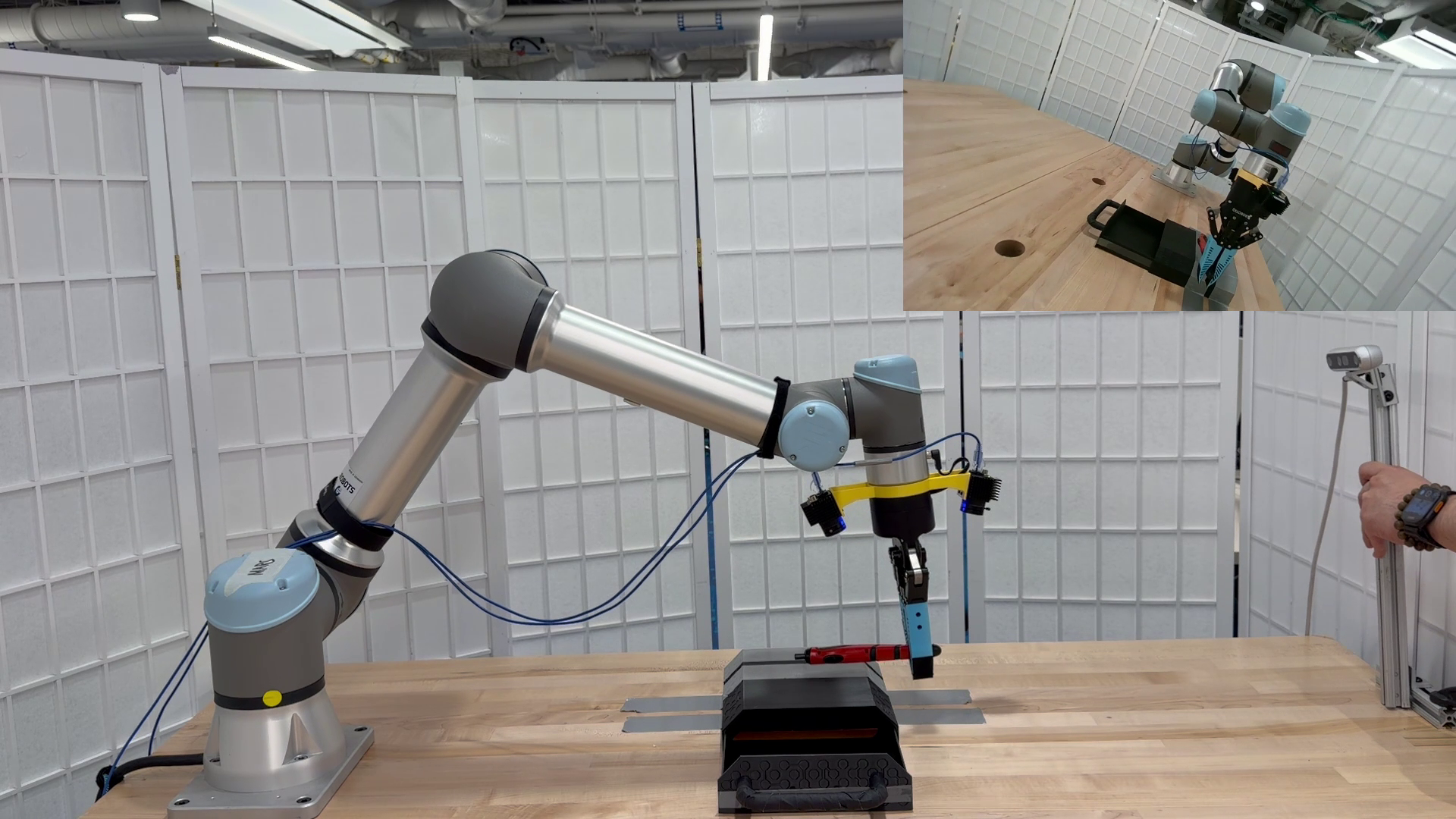}
  \end{subfigure}\hspace{1pt}%
  \begin{subfigure}[t]{0.195\textwidth}
    \includegraphics[width=\linewidth]{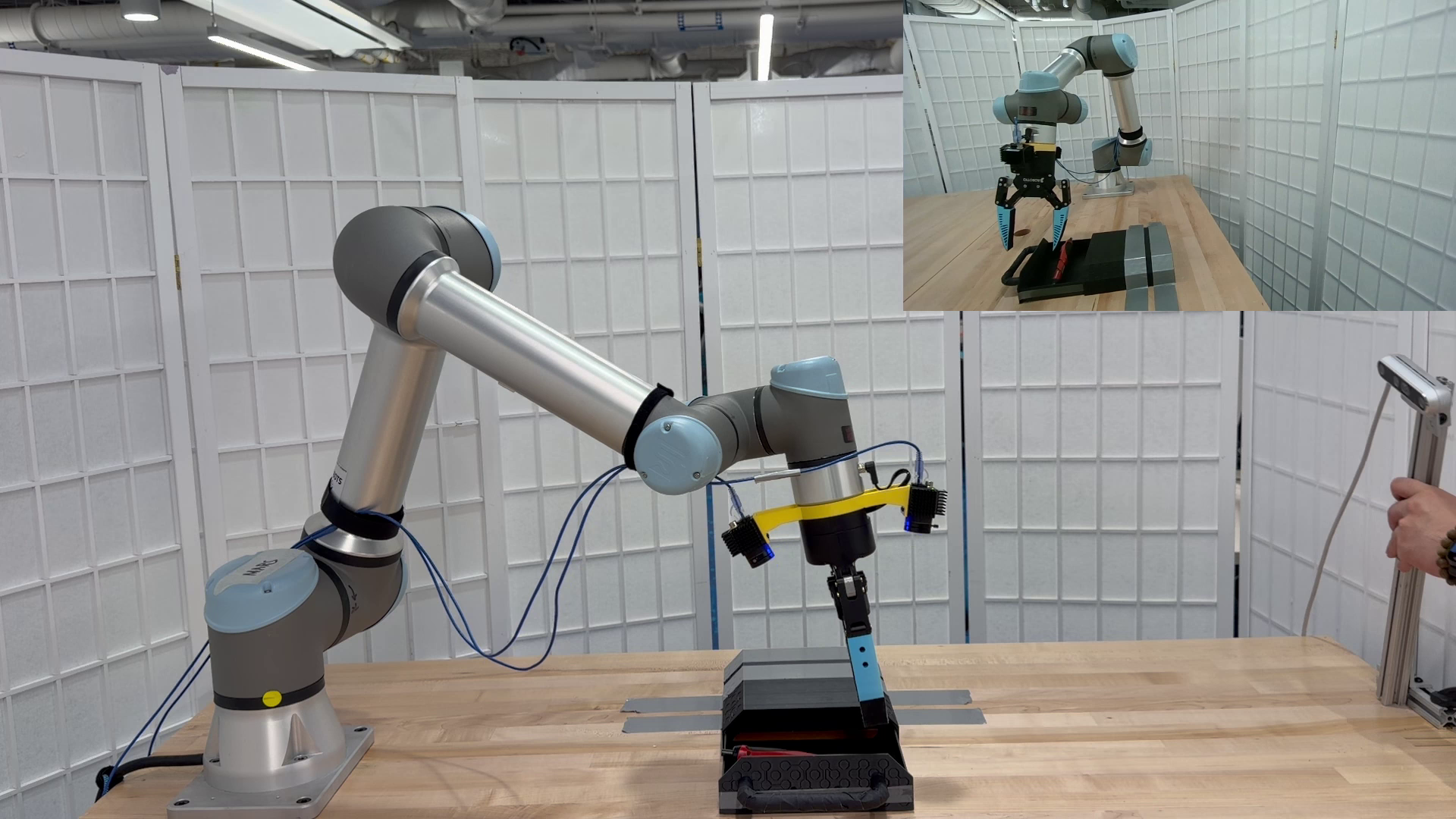}
  \end{subfigure}

  \caption{\textbf{Agent-view camera perturbations during evaluation.} Each row shows five representative agent-view frames captured with the side-view camera randomly perturbed (\textbf{A small figure embedded in the top right }), for the \textbf{Plug Flower} task (top) and \textbf{Put in Drawer} task (bottom). The in-hand cameras are rigidly mounted to the end-effector and therefore cannot be perturbed independently. $\ours$ remains stable across these viewpoint changes, while the baseline fails to complete the tasks.}
  \label{fig:real-shake}
\end{figure}

\textbf{Side-view camera perturbations.}
To evaluate robustness to camera viewpoint changes, we randomly perturb the agent-view camera during inference. We perturb only the agent-view camera because the in-hand cameras are rigidly mounted to the end-effector and therefore cannot be shifted independently of the robot. Figure~\ref{fig:real-shake} shows six representative perturbed agent-view frames for the \textbf{Plug Flower} and \textbf{Put in Drawer} tasks, 
illustrating the range of viewpoint changes the policy must handle.

\newpage
\subsection{Real-World Experiments with Reduced Demonstrations}
\label{reduced_demo}

\begin{wraptable}[7]{r}{0.42\textwidth}
  \vspace{-1em}
  \setlength\tabcolsep{4pt}
  \scriptsize
  \centering
  \begin{tabular}{@{}lccc@{}}
    \toprule
    Task & \#Demo & $\ours$ & DP \\
    \midrule
    \multirow{2}{*}{Prepare Fruit Plate} & 40  & 55 & 20 \\
                                          & 70  & 85 & 60 \\
    \midrule
    \multirow{2}{*}{Make Toast}           & 70  & 80 & 55 \\
                                          & 100 & 85 & 60 \\
    \bottomrule
  \end{tabular}
  \caption{\textbf{Sample efficiency.} Real-world success rates (\%) over 20 trials per setting at different demonstration budgets.}
  \label{table:reduced-demo}
  \vspace{-1em}
\end{wraptable}
To assess sample efficiency, we evaluate $\ours$ against Diffusion Policy (DP)~\cite{chi2023diffusion} under reduced demonstration budgets. We test on {Prepare Fruit Plate} with $40$ and $70$ demonstrations and on {Make Toast} with $70$ and $100$ demonstrations. We report completion rates over 20 trials per task, counting each trial as either a success or a failure with no partial credit.

\subsection{Simulated Task Description}
\label{appendix-sim}

We evaluate on 10 tasks from MimicGen~\cite{mandlekar2023mimicgen}. The full task description used for each task is provided below.
\textbf{stack-three-d1}: stack three blocks on top of each other.
\textbf{pick-place-d0}: pick up three objects and place them into the corresponding slots sequentially.
\textbf{kitchen-d1}: toggle the stove, place the pan on the stove, put the item inside the pan, then place it on the red plate, and finally turn off the stove.
\textbf{hammer-cleanup-d1}: open the drawer, pick up the hammer, place it inside, and close the drawer.
\textbf{mug-cleanup-d1}: open the drawer, place the mug inside, and close the drawer.
\textbf{coffee-d2}: pick up the coffee capsule, insert it into the cup, and close the cap.
\textbf{coffee-preparation-d1}: grasp the coffee mug and place it on the coffee machine, then open the drawer, take the coffee capsule, and place it into the coffee machine.
\textbf{nut-assembly-d0}: assemble a nut onto the same-colored bar.
\textbf{square-d2}: assemble the square into the slot.
\textbf{threading-d2}: thread the needle into the slot.

\begin{figure*}[b]
    \centering
    \includegraphics[width = 0.99\textwidth]{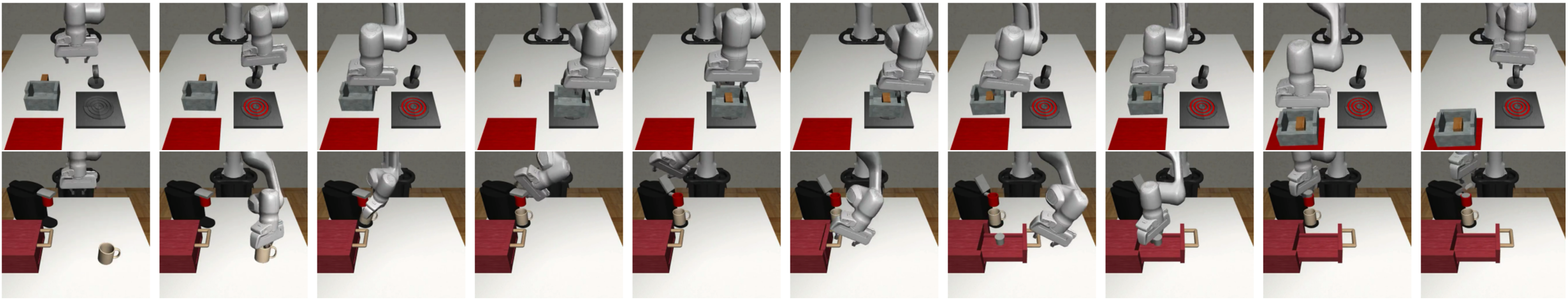}
    \caption{\textbf{Long-horizon task illustration.} The top row shows \textbf{kitchen-d1}, and the bottom row shows \textbf{coffee-preparation-d1}.
    \vspace{-0.1cm}}
    \label{fig:long-horizon-task}
    \vspace{-0.2cm}
\end{figure*}

\begin{wrapfigure}[9]{r}{0.5\textwidth}
  \vspace{-1em}
  \centering
  \includegraphics[width=0.48\textwidth]{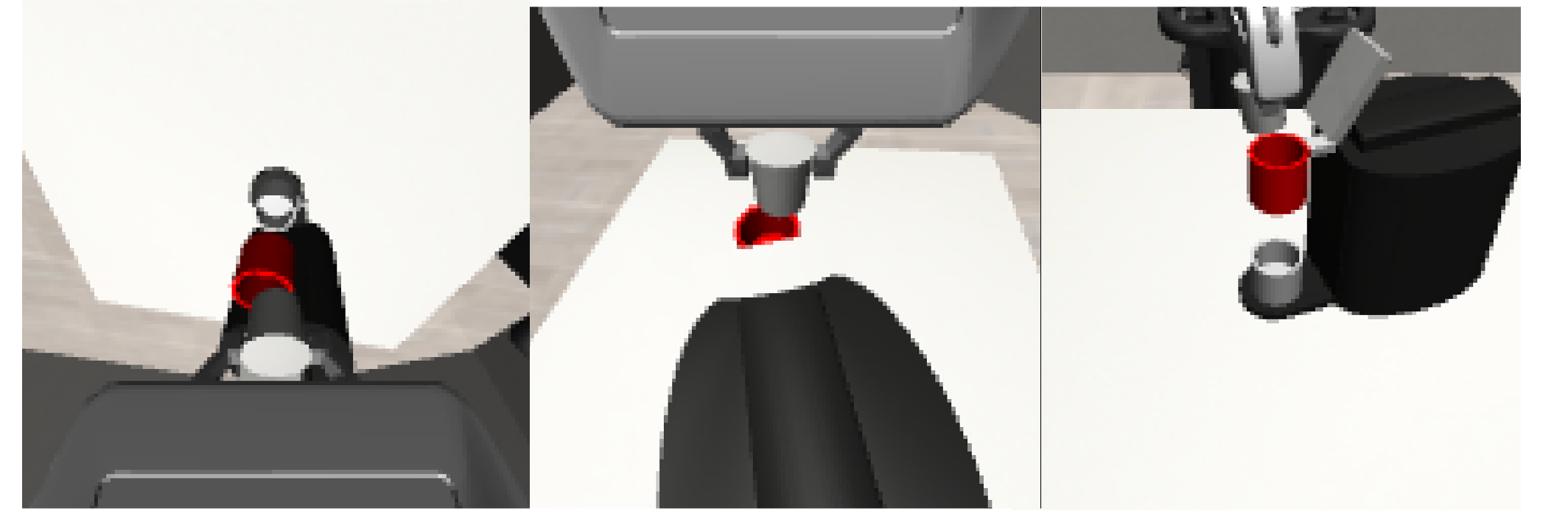}
  \caption{\textbf{Observed images in simulation.} From left to right: top in-hand view, bottom in-hand view, and agent view. Each is a $128 \times 128$ RGB image.}
  \label{fig:sim-img}
  \vspace{-1em}
\end{wrapfigure}

\textbf{Task categories.} We organize the tasks along four overlapping axes used throughout the paper. By precision level, three tasks involve pick-and-place precision (stack-three-d1, kitchen-d1, pick-place-d0); four involve articulated-object manipulation requiring higher precision (hammer-cleanup-d1, mug-cleanup-d1, coffee-d2, coffee-preparation-d1); and three involve high-precision insertion (nut-assembly-d0, square-d2, threading-d2). By spatial variation, the d0/d1/d2 suffixes denote light, moderate, and substantial variation in the test distribution; we use the largest variant provided by MimicGen~\cite{mandlekar2023mimicgen} for each task. Three tasks are long-horizon, involving multiple sequential 
\begin{wrapfigure}{r}{0.65\textwidth}
  \vspace{-0em}
  \centering
  \includegraphics[width=0.65\textwidth]{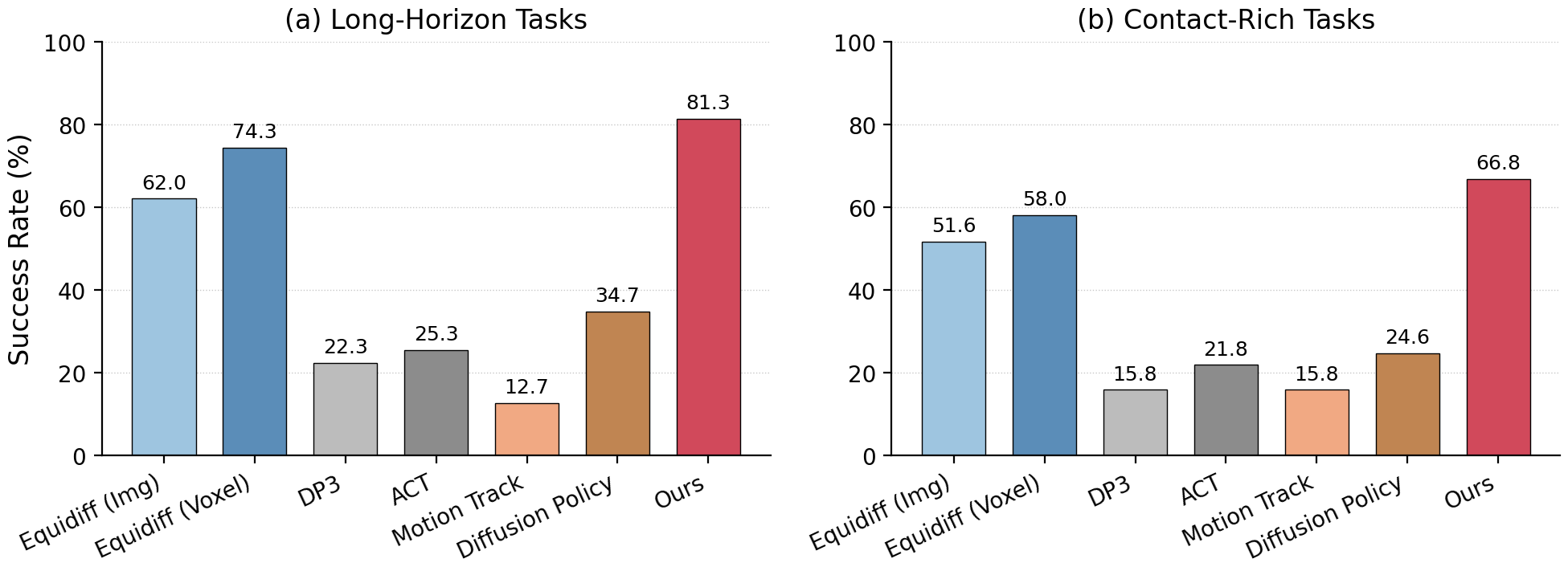}
  \caption{Performance Comparisons on Different Types of Task.}
  \label{fig:long-horizon-contact-rich}
  \vspace{-1em}
\end{wrapfigure}
object interactions (kitchen-d1, pick-place-d0, coffee-preparation-d1); see Figure~\ref{fig:long-horizon-task} for example rollouts. Five tasks are contact-rich, involving precise insertion (coffee-d2, coffee-preparation-d1, nut-assembly-d0, square-d2, threading-d2). A per-category comparison against baselines is provided in Figure~\ref{fig:long-horizon-contact-rich}.
\textbf{Embodiment.} We use the Panda arm for most tasks, except \textit{pick-place-d0} and \textit{nut-assembly-d0}, which use the Sawyer arm following the MimicGen defaults. Observations from the two in-hand cameras and one agent-view camera used in simulation are shown in Figure~\ref{fig:sim-img}.

\subsection{Baseline Description}
\label{appendix-baseline}

We compare $\ours$ against a diverse set of baselines spanning different input modalities and architectural choices. Results for Equidiff~\cite{wang2024equivariant}, ACT~\cite{zhao2023learning}, and DP3~\cite{ze20243d} are taken from~\cite{wang2024equivariant}. \textbf{Diffusion Policy} and \textbf{Motion Track} use the same image-based setup as $\ours$ (two in-hand cameras and one agent view). \textbf{DP3} and \textbf{EquiDiff (Voxel)} use 3D perception, which provides geometric information unavailable from 2D images alone; following the setup in~\cite{wang2024equivariant}, both take four depth cameras paired with one in-hand view as input, from which point clouds (for DP3) and voxels (for EquiDiff) are extracted. \textbf{EquiDiff (Img)} and \textbf{ACT} share the same architecture but use only a single in-hand camera. Together, these baselines cover a wide range of observation modalities (2D images, point clouds, voxels) and policy classes (diffusion, equivariant, Transformer-based), providing a thorough comparison.

\subsection{Implementation Details}
\label{implementation-details}

\textbf{Action representation.} We canonicalize the 2D image-coordinate trajectories by subtracting the image-space gripper-center keypoint. Since an image translation shifts both the gripper center and the image action chunk identically, this operation induces translation invariance in the canonicalized representation, analogous to relative-action designs that express actions with respect to the current end-effector pose.

\textbf{Handling out-of-plane motion.} Because we project the 3D action chunk onto the image plane defined by the current end-effector pose, future motions can occasionally cross this plane. In such cases, especially when the normalization depth is small, the resulting image coordinates may become excessively large. We clean the training data by setting a large threshold: image coordinates whose magnitude exceeds the threshold are replaced with the previous action. This can be interpreted as holding the action fixed when the next sample becomes numerically unstable.

\textbf{Visual encoders.} The left branches of X-Net take the two in-hand images and one agent-view image as input. Each is processed by a ResBlock-based convolutional encoder with four ResBlocks, kernel size 3, and $\text{MaxPool}(2,2)$ downsampling layers that trade spatial resolution for channel capacity. The in-hand encoder uses hidden channel dimensions of $\{64, 128, 256, 512, 1024\}$; the agent-view encoder is identical except for the final stage, with $\{64, 128, 256, 512, 512\}$. The two in-hand views share weights, while the agent-view encoder is separate. After encoding, the in-hand feature map is split along the channel dimension and flattened spatially, yielding $512$-dimensional visual tokens for each of the two in-hand cameras and the agent view.

\textbf{Multi-view Transformer.} The center of X-Net is a multi-view Transformer that mediates communication between the in-hand and agent-view features. It consists of six attention layers (two in-image, four cross-image), with $8$ heads and a per-head dimension of $64$. When available, language and proprioceptive tokens are injected through the cross-attention mechanism. Only the $512$-dimensional tokens corresponding to the two in-hand cameras are retained at the output and concatenated with the second channel split from the visual encoder.

\textbf{Diffusion head.} The right branches of X-Net regress two sets of continuous keypoint trajectories independently from the two in-hand feature streams, with $\SE(2)$ augmentation applied per stream — encouraging the model to learn an $n$-equivariant function across views. Action generation is performed by a DiT-X diffusion head that takes noisy actions and in-hand visual features as input and predicts the noise component. The Transformer backbone has $12$ layers, $8$ heads per layer, and an embedding dimension of $768$. \textbf{Crucially, the diffusion head is shared across the two in-hand views}, which is the key design choice enabling permutation equivariance.

\textbf{Parameter count.} The diffusion head contains approximately $1.8 \times 10^8$ parameters; the image encoders contain approximately $5.7 \times 10^7$ parameters in total.

\subsection{Training and Evaluation Details}
\label{appendix-training-details}

\textbf{Training.} We train the model with an initial learning rate of $10^{-4}$, decayed via a cosine schedule. For simulated experiments, we use DDPM with 100 denoising steps; for real-world experiments, we use DDIM with 16 denoising steps for faster inference. The model is trained with a batch size of $256$ for $800$ epochs.

\textbf{Data augmentation.} We apply rotation and translation transformations independently to each camera view and its corresponding image-space action. Rotations are sampled uniformly from $[-30^\circ, +30^\circ]$ per camera: rotated images are obtained via bilinear interpolation, while rotated keypoints are computed analytically by applying the corresponding rotation matrix in continuous space. Translations shift each image by up to $\pm h/8$ and $\pm w/8$, where $h$ and $w$ are the image height and width. This extensive augmentation encourages the model to reason about keypoint trajectories in the image plane and to capture local geometric features that determine the correct action.

\textbf{Simulation evaluation.} For each task, we evaluate on $50$ unseen test episodes with spatial variations and report the best performance observed during training, averaged over two seeds.

\textbf{Real-world deployment and inference speed.} The inference time of $\ours$ from observation to final denoised action is approximately $58$\,ms on an NVIDIA RTX 4090 GPU. We deploy the policy at $10$\,Hz with an action horizon of $10$, which is interpolated by the robot controller into high-frequency control signals. To compensate for system latency, we discard the first $n$ actions of each predicted action chunk before execution.

\textbf{Real-world failure case analysis.}
\label{appendix-failure}
The main failure mode of $\ours$ is imprecise grasping and insertion, occurring primarily under reduced demonstration budgets and out-of-distribution initial configurations. Take make-toast as an example: even when the model successfully inserts the first slice, small misalignments when inserting the second can cause the task to fail, as it tolerates only about a 5-degree rotation offset.

\subsection{Pseudocode}
  \label{appendix:pseudocode}

  Algorithms~\ref{alg:p2a-train} and~\ref{alg:p2a-infer} summarize the training
  and inference of \ours{} in the notation of Section~\ref{3d_2d_transformation}.
  The observation $\mathcal{O}{=}(o_1,o_2,o_3)$ stacks the two in-hand views
  $o_1,o_2$ (camera matrices $c_1,c_2$) and the head view $o_3$ (context only).
  $\mathcal{M}$ maps a pose to its $m{=}4$ gripper keypoints
  $p_i^{1:4}{:=}(p_i^1,\dots,p_i^4)$ and back, keeping the width $w$ as a separate
  scalar; $\mathcal{P},\mathcal{T}$ are projection and triangulation;
  $A_{\mathrm{pix}}^k$ is the image action chunk for in-hand view $k$;
  $\epsilon_\theta$ is the Diffusion X-Net noise predictor, \emph{shared} across
  the two in-hand views; $g_k\!\in\!\SE(2)$ are per-camera augmentations; $s$ is
  the image size, $w_{\max}$ the maximum gripper width, $\rho$ a large stability threshold
  ($\rho{\approx}1.5s$, well outside the image), so coordinates stay
  \emph{unbounded} and may leave the frame --- only plane-crossing blow-ups with
  $|\mathrm{pix}_{ijk}|{>}\rho$ are held at the last valid value; $\bar\alpha_\tau$ the
  diffusion noise schedule, and $T_{\mathrm{ee}}$ the current end-effector pose,
  which expresses the in-hand cameras $c_k$ relative to the gripper
  (canonicalization). For brevity each line applies over all horizon steps
  $i{\le}h$, keypoints $j{\le}m$, and in-hand views $k\!\in\!\{1,2\}$. In training,
  lines~1--4 build the image-action labels, 5--7 apply equivariant augmentation,
  and 8--12 are the standard diffusion objective.

  \vspace{4pt}
  \noindent
  \begin{minipage}[t]{0.48\textwidth}
  \begin{algorithm}[H]
  \footnotesize
  \caption{\ours{} -- Training step}
  \label{alg:p2a-train}
  \begin{algorithmic}[1]
  \Require demo $(\mathcal{O},\mathcal{A}{=}(a_i)_{i=1}^{h})$;\ $\epsilon_\theta$;\ rate $\eta$
  \State $c_k \gets T_{\mathrm{ee}}\,c_k$
  \State $p_i^{1:4},\,w_i \gets \mathcal{M}(a_i)$
  \State $\mathrm{pix}_{ijk}\gets \mathcal{P}(p_i^j;\,c_k)$
  \State hold last valid $(\mathrm{pix}_{ijk},w_i)$ once $|\mathrm{pix}_{ijk}|{>}\rho$
  \State sample $g_1,g_2,g_3\sim\SE(2)$
  \State $(o_k,\mathrm{pix}_{ijk})\gets g_k\!\cdot\!(o_k,\mathrm{pix}_{ijk})$
  \State $o_3\gets g_3\!\cdot\! o_3$
  \State $A_{\mathrm{pix}}^k\gets(\mathrm{pix}_{ijk}/s,\ w_i/w_{\max})$
  \State $\epsilon\sim\mathcal{N}(0,I),\ \tau\sim\mathcal{U}$
  \State $A_\tau\gets\sqrt{\bar\alpha_\tau}\,A_{\mathrm{pix}}+\sqrt{1{-}\bar\alpha_\tau}\,\epsilon$
  \State $\mathcal{L}\gets\|\epsilon_\theta(\mathcal{O},A_\tau,\tau)-\epsilon\|^2$
  \State $\theta\gets\theta-\eta\,\nabla_\theta\mathcal{L}$
  \end{algorithmic}
  \end{algorithm}
  \end{minipage}
  \hfill
  \begin{minipage}[t]{0.48\textwidth}
  \begin{algorithm}[H]
  \footnotesize
  \caption{\ours{} -- Inference}
  \label{alg:p2a-infer}
  \begin{algorithmic}[1]
  \Require obs $\mathcal{O}$;\ $\epsilon_\theta$;\ $\mathcal{T}$;\ $K$ steps
  \State $c_k \gets T_{\mathrm{ee}}\,c_k$
  \State $A\sim\mathcal{N}(0,I)$
  \For{$\tau=K$ \textbf{downto} $1$}
    \State $A\gets \textsc{Denoise}\big(A,\,\epsilon_\theta(\mathcal{O},A,\tau)\big)$
  \EndFor
  \State $(\mathrm{pix}_{ijk},w_i^k)\gets A$
  \State $\mathrm{pix}_{ijk}\gets s\,\mathrm{pix}_{ijk}$
  \State $p_i^j\gets \mathcal{T}(\mathrm{pix}_{ij1},\mathrm{pix}_{ij2};\,c_1,c_2)$
  \State $\bar w_i\gets\tfrac12(w_i^1{+}w_i^2)\,w_{\max}$
  \State $a_i\gets \mathcal{M}^{-1}(p_i^{1:4},\bar w_i)$
  \State \Return $\mathcal{A}=(a_1,\dots,a_h)$
  \State execute first $n_a$ steps, then re-plan
  \end{algorithmic}
  \end{algorithm}
  \end{minipage}

\end{document}